\documentclass{article}

\usepackage{arxiv} 
\usepackage[utf8]{inputenc} 
\usepackage[T1]{fontenc}    
\usepackage[numbers]{natbib} 

\usepackage{graphicx}       
\usepackage{booktabs}       
\usepackage{multirow}       
\usepackage{amsfonts}       
\usepackage{amsmath, amssymb} 
\usepackage{nicefrac}       
\usepackage{enumitem}       
\usepackage{microtype}      
\usepackage{textcomp}       
\usepackage{xcolor}         
\usepackage{url}            
\usepackage{doi}            
\usepackage{fancyhdr}       
\usepackage{enumitem}       

\usepackage{hyperref}
\hypersetup{
    colorlinks=true,
    linkcolor=blue,
    citecolor=blue,
    filecolor=magenta,      
    urlcolor=cyan,
    pdftitle={Transformer-Based Approach to Optimal Sensor Placement},
}

\newcommand{\orcidicon}[1]{\href{https://orcid.org/#1}{\includegraphics[scale=0.06]{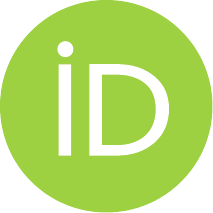}}}

\begin{document}

\title{Transformer-Based Approach to Optimal Sensor Placement for Structural Health Monitoring of Probe Cards}

\date{}

\author{
    \orcidicon{0000-0002-7668-3413}\hspace{1mm}Mehdi Bejani \\
    Dept. of Civil and Environmental Engineering \\
    Politecnico di Milano, Italy \\
    Technoprobe, Cernusco Lombardone, Italy \\
    \texttt{mehdi.bejani@polimi.it} \\
    \And
    \orcidicon{0009-0005-0553-4348}\hspace{1mm}Marco Mauri \\
    Technoprobe \\
    Cernusco Lombardone, Italy \\
    \texttt{marco.mauri@technoprobe.com} \\
    \And
    Daniele Acconcia \\
    Technoprobe \\
    Cernusco Lombardone, Italy \\
    \texttt{daniele.acconcia@technoprobe.com} \\
    \And
    Simone Todaro \\
    Technoprobe \\
    Cernusco Lombardone, Italy \\
    \texttt{simone.todaro@technoprobe.com} \\
    \And
    \orcidicon{0000-0001-5111-9800}\hspace{1mm}Stefano Mariani \\
    Dept. of Civil and Environmental Engineering \\
    Politecnico di Milano, Italy \\
    \texttt{stefano.mariani@polimi.it} \\
}

\maketitle

\begin{abstract}
This paper presents an innovative Transformer-based deep learning strategy for optimizing the placement of sensors aiming at structural health monitoring of semiconductor probe cards. Failures in probe cards, including substrate cracks and loosened screws, would critically affect semiconductor manufacturing yield and reliability. Some failure modes could be detected by equipping a probe card with adequate sensors. Frequency response functions from simulated failure scenarios are adopted within a finite element model of a probe card. A comprehensive dataset, enriched by physics-informed scenario expansion and physics-aware statistical data augmentation, is exploited to train a hybrid Convolutional Neural Network and Transformer model. The model achieves high accuracy (99.83\%) in classifying the probe card health states (baseline, loose screw, crack) and an excellent crack detection recall (99.73\%). Model robustness is confirmed through a rigorous framework of 3 repetitions of 10-fold stratified cross-validation. The attention mechanism also pinpoints critical sensor locations: an analysis of the attention weights offers actionable insights for designing efficient, cost-effective monitoring systems by optimizing sensor configurations. This research highlights the capability of attention-based deep learning to advance proactive maintenance, enhancing operational reliability and yield in semiconductor manufacturing.
\end{abstract}

\keywords{Optimal Sensor Placement \and Neural Networks \and Transformer \and Attention Mechanism \and Probe Card \and Failure Mode Analysis \and Real-time Monitoring \and Sensor Networks \and Structural Health Monitoring \and Electrical Wafer Sort}

\section{Introduction}
\label{sec:introduction}
In the intricate landscape of semiconductor manufacturing, where precision and reliability are paramount, probe cards (PCs) stand as indispensable components. These critical intermediary tools bridge the gap between silicon wafers and automated test equipment (ATE), playing a crucial role in identifying and preventing defective integrated circuit (IC) dies to next move on to the resource-intensive packaging stage, thereby mitigating consumer risks and minimizing potential economic losses \citep{ref1}. As the primary interface for the electrical testing of ICs fabricated on wafers, the integrity and functionality of PCs directly impact the quality assurance process. Failures in these sophisticated devices, which can manifest as microscopic cracks within the substrate or the loosening of screws that affects the crucial contact integrity, can lead to a cascade of detrimental outcomes. These include the generation of inaccurate test results, the erroneous rejection of functional dies, a significant reduction in the overall manufacturing yield, and the incurrence of substantial costs associated with unscheduled equipment downtime should these issues remain undetected.

As the semiconductor industry continues its relentless pursuit of miniaturization and increased complexity in IC design, the demands placed on PC technology have intensified. These advancements necessitate PCs capable of enabling reliable, repeatable, and consistently accurate electrical contacts at ever-decreasing scales. Consequently, contemporary PC technologies must facilitate exceptionally precise measurements, leading to a profound influence on both the yield and the long-term reliability of semiconductor products \citep{ref2}. Modern PCs are engineered with highly intricate architectures, often comprising tens of thousands of pins. This high level of integration introduces substantial complexity in terms of maintenance, diagnostics, and fault isolation \citep{ref3, ref4}. The occurrence of critical failure modes within these systems can result in operational disruptions, decreased testing throughput, and overall degradation in manufacturing efficiency and customer satisfaction. These failures typically arise from a combination of mechanical fatigue, electrical performance deterioration, and exposure to adverse environmental conditions. Therefore, the ability to anticipate and accurately forecast such failure mechanisms is essential for maintaining the operational reliability and productivity of PC systems.

Traditional methodologies employed for the detection of PC failures rely on manual visual inspections or electrical tests, which may only succeed in identifying problems once a significant level of degradation has already occurred. Furthermore, such methods often necessitate the interruption of the testing process, leading to inefficiencies in the production workflow. In practice, maintenance engineers frequently depend on their accumulated domain-specific knowledge and engage in time-consuming trial-and-error procedures to diagnose faults and undertake troubleshooting efforts \citep{ref1}. Given the continuous shrinking of semiconductor feature sizes, the processes of diagnosing and troubleshooting issues associated with PCs have become increasingly intricate and time-consuming. This growing complexity underscores the limitations of relying solely on human expertise and reactive maintenance strategies.

Recent advancements in semiconductor manufacturing, coupled with the continued escalation in circuit complexity as predicted by Moore's law, have significantly intensified the challenges associated with fault diagnosis in Electrical Wafer Sort (EWS) systems \citep{ref3, ref5}. In response, artificial intelligence (AI) methodologies have gained widespread traction within the semiconductor industry as effective tools for addressing these diagnostic challenges. For example, modular neural networks and Bayesian inference models have been employed for fault diagnosis in advanced semiconductor equipment \citep{ref6, ref7}, while Support Vector Machines (SVMs) have demonstrated utility in anomaly detection and fault classification tasks \citep{ref8}. Additionally, autoencoder-based frameworks, combined with event log data, have introduced innovative strategies for early fault detection in the EWS processes \citep{ref9}. AI techniques have further been extended to tasks such as PC lifetime prediction and optimization of replacement schedules \citep{ref10}, as well as differentiating between defects caused by testing apparatus and those inherent to the fabrication process \citep{ref11}.

In response to the shortcomings of traditional failure detection methods, Structural Health Monitoring (SHM) techniques have emerged as a promising alternative. By leveraging sensor data, such as measurements of vibrations, SHM offers the potential for continuous and non-intrusive monitoring of the operational health of probe cards. The primary aim of this research is to establish a comprehensive SHM framework tailored to the detection and management of failure modes in PCs employed in EWS applications. Through the systematic analysis of historical failure data collected over a statistically significant duration, this study identifies and prioritizes critical failure mechanisms. Among these, mechanical failures such as cracks in ceramic substrates and loosened fastening components emerge as particularly detrimental, even if rare, due to their pronounced impact on performance and the inherent difficulty in early-stage detection. Such failures can lead to irreversible damage, prolonged downtime, and diminished operational efficiency. To address these issues, the proposed approach integrates a sensor network with predictive modeling techniques to enable proactive failure management. Specifically, this study emphasizes the Optimal Sensor Placement (OSP) using Deep Learning (DL) methodologies, with the objective of maximizing the informational value of the sensor network within the spatial constraints of the PC architecture.

Among the various types of sensor data, Frequency Response Functions (FRFs), which are mathematically derived from the dynamic responses of a structure, exhibit a particular sensitivity to alterations in the structural integrity and have demonstrated their utility in the identification of damage and mechanical failures \citep{ref12}. An FRF essentially provides a numerical representation of the relationship between the input and output of a system in the frequency domain, revealing critical information about its resonant frequencies, damping characteristics, and mode shapes, all of which are susceptible to changes caused by structural impairments \citep{ref13}. Damage to a structure, whether it occurs in the form of compromised connections or material degradation, inevitably leads to modifications in its inherent dynamic properties, which are subsequently reflected in the corresponding FRF data. By analyzing FRF data from a potentially damaged structure and comparing it to that of an undamaged baseline, changes such as shifts in the natural frequencies can be observed, indicating the presence, and potentially the location and severity, of the damage \citep{ref12}. This capability positions FRF analysis as a valuable tool for the non-destructive and potentially real-time assessment of PC health, offering notable advantages over conventional methods.

Despite significant strides in AI applications for semiconductor manufacturing, a critical research gap remains in the development of AI-driven systems for the comprehensive online monitoring and management of PC health. By addressing this gap, the present study proposes an intelligent sensor network framework specifically designed for SHM of PCs, with a particular emphasis on OSP strategies. Prior research on OSP has predominantly employed conventional approaches, including the Fisher Information Matrix and related criteria \citep{ref14, ref15, ref16}, information entropy-based methods \citep{ref17, ref18}, the value of information framework \citep{ref19, ref20}, and Modal Assurance Criterion based Methods \citep{ref21, ref22, ref23, ref24}. In contrast, this study introduces a novel DL-based OSP methodology tailored to the unique challenges of PC monitoring. Building upon established AI frameworks \citep{ref25}, we integrate attention mechanisms within DL architectures to identify sensor locations that yield the highest diagnostic value. Our approach leverages simulation data generated through ANSYS Mechanical R2 2024 \citep{ref26}, encompassing diverse failure scenarios. By analyzing attention scores produced by the network, we identify sensor configurations that maximize information gain while minimizing redundancy. This AI-assisted strategy enhances monitoring precision, reduces operational costs, and significantly improves the reliability and maintainability of PC systems.

The analysis of the complex and high-dimensional FRF data necessitates the application of advanced signal processing and pattern recognition techniques. In this context, DL has witnessed a significant rise as a powerful tool capable of automatically learning intricate features directly from raw sensor data, demonstrating remarkable success across a diverse range of SHM applications \citep{ref27, ref28, ref29}. Specifically, Convolutional Neural Networks (CNNs) have proven to be particularly well-suited for processing time-series or spectral data, such as FRFs, due to their inherent ability to effectively extract localized patterns \citep{ref30, ref31}. Furthermore, Recurrent Neural Networks (RNNs), and more recently, Transformer networks, have demonstrated their strength in capturing sequential dependencies and global relationships within complex datasets \citep{ref28, ref31, ref32}. The capacity of DL methodologies to automate the intricate analysis of sensor data, including FRFs, addresses the limitations often associated with traditional signal processing techniques, which frequently require extensive manual intervention for feature engineering.

This paper thus provides key contributions as follows:
\begin{itemize}
    \item Development of a hybrid CNN-Transformer model, leveraging attention mechanisms for PC failure detection and interpretable OSP.
    \item Creation of an expanded and diverse dataset of FRF responses, through physics-informed scenario expansion and physics-aware data augmentation based on Finite Element (FE) simulations.
    \item Implementation of a rigorous methodological framework, including strict data handling, to prevent leakage and robust cross-validation for reliable model evaluation.
    \item Demonstration of the model high classification performance and its ability to identify critical sensor locations, offering a data-driven approach to SHM system design for PCs.
\end{itemize}
The remainder of this paper is structured as follows: Section \ref{sec:related_work} reviews related work on DL in SHM, particularly for PCs, emphasizing the role of Transformer networks. Section \ref{sec:methodology} details the methodology: dataset generation, preprocessing, augmentation, the CNN-Transformer model architecture, and evaluation protocols. Section \ref{sec:results} collects experimental results: performance metrics, confusion matrices, learning curves, and sensor importance profiles from attention weights. Section \ref{sec:discussion} discusses findings, compares with prior research, highlights OSP implications, and outlines limitations and future work. Finally, Section \ref{sec:conclusion} concludes with a summary of key contributions and their significance.

\section{Related Work}
\label{sec:related_work}
The application of machine learning, and particularly of DL, to the field of SHM and fault detection has experienced a rapid growth in recent years, see e.g. \citep{ref33, ref34, ref35}. This section aims to provide a concise overview of relevant studies that have explored the use of sensor data in industrial contexts, with a specific focus on DL techniques and the increasing prominence of Transformer networks. Early investigations in SHM often relied on the integration of traditional signal processing techniques with classical machine learning algorithms, such as SVMs \citep{ref36, ref37} or artificial neural networks \citep{ref38, ref39} and k-means clustering \citep{ref40, ref41}. However, these methods typically required a significant domain-specific expertise for the crucial task of feature engineering. DL methodologies have effectively addressed this limitation by offering the capability to automatically learn hierarchical representations of features directly from raw sensor data \citep{ref27, ref42, ref43, ref44, ref45, ref46}. This shift towards automated feature extraction has been a key driver in the increased adoption of DL in various SHM applications.

Within the context of SHM, CNNs have emerged as a highly effective architecture for the analysis of data in either time or frequency domain. Their success has been demonstrated in various applications, including vibration analysis for the diagnosis of faults in machinery \citep{ref30}, acoustic signal processing for the detection of anomalies \citep{ref47}, and FRF analysis for the identification of structural damage \citep{ref12}. The inherent ability of CNNs to capture local patterns through the application of convolutional filters makes them particularly well-suited for extracting meaningful information from sensor signals. Previous work has successfully applied a CNN-based model to the analysis of FRF data obtained from PCs, demonstrating the feasibility and potential of this approach for detecting failures in these components.

Although CNNs have proven to be exceptionally effective in processing data from individual sensor streams, they are not inherently designed to model the complex interactions that may exist between different sensors or to capture long-range dependencies that span across the entire set of sensors simultaneously; some results have anyway been obtained, see e.g. \cite{Rosafalco2020,ref25}. RNNs, particularly long short-term memory networks and gated recurrent units, are capable of processing sequential data from sensors. However, effectively handling a fixed set of independent sensor streams concurrently using standard RNN architectures can be challenging without specific architectural modifications \citep{ref31, ref32}.

Transformer networks, which were initially developed for applications in natural language processing \citep{ref48, ref49}, have recently achieved remarkable success in a wide range of domains, including computer vision and time-series forecasting \citep{ref30, ref50, ref51}. The core mechanism underlying the power of Transformers is self-attention, which allows the model to dynamically weigh the importance of different parts of the input sequence (or, in the context of sensor data, different sensor features) when making a prediction. This inherent ability to model global dependencies and intricate relationships within data makes Transformer networks particularly appealing for tasks that involve multiple, interacting sensors. Within SHM, Transformers have been already applied for sequence modeling of multivariate sensor data and for anomaly detection \citep{ref50, ref52}.

A significant advantage of the attention mechanism in Transformers is its interpretability. By examining attention weights, it is possible to infer which input parts (e.g., which sensors) were most influential in the model decision, a valuable feature for PC monitoring as the identification of critical sensor locations can inform OSP strategies \citep{ref25, ref28, ref53}. While Transformer networks have been explored for general multivariate time series analysis, their specific application to PC failure detection, leveraging attention for interpretable OSP within a rigorous cross-validation framework, represents a novel contribution. The combination of CNNs local feature extraction with Transformers global interaction modeling and interpretability offers a promising avenue for improving performance and understanding in this critical application.

This work builds upon a previous CNN-based study \citep{ref53} by:
\begin{itemize}
    \item Utilizing a significantly expanded and more diverse dataset, generated through systematic physics-informed scenario variations.
    \item Implementing enhanced methodological rigor, particularly in data handling to prevent leakage and in employing a robust cross-validation scheme.
    \item Introducing a more complex hybrid CNN-Transformer model architecture designed to capture both local sensor features and global inter-sensor dependencies.
    \item Placing a strong and reliability emphasis on the interpretable analysis of sensor importance via attention mechanisms for OSP.
\end{itemize}
Previous research on DL for PC failure detection using FRF data had limitations such as dataset size, potential data leakage from augmentation strategies applied before data splitting, and a lack of detailed sensor importance analysis. This paper directly addresses these shortcomings. We introduce a robust framework leveraging an expanded dataset, a strict data splitting strategy preceding augmentation, and a 10-fold stratified cross-validation approach with multiple seeds. The core is a novel CNN-Transformer model that enhances classification and provides sensor importance insights via attention weights, demonstrating superior performance and robustness. This aligns with the growing trend of attention mechanisms for interpretable sensor data analysis \citep{ref25, ref54, ref55}.

\section{Methodology}
\label{sec:methodology}
This section details the methodology employed in this study. A description of the problem formulation and of the proposed SHM solution is given first, followed by an in-depth explanation of the digital shadow used in the FE simulations. Subsequently, the challenges related to data imbalance and the strategies for physics-informed scenario expansion and data augmentation are discussed. The architecture of the proposed hybrid CNN-Transformer model is then presented. Finally, the cross-validation framework and the techniques used for sensor importance analysis are outlined.

\subsection{Problem Formulation and Proposed SHM Solution}
\label{subsec:problem_formulation}
The central aim of this study is the development of an AI-assisted SHM framework for real-time monitoring and management of PCs, with the objective of enhancing their operational reliability and process efficiency in wafer-level electrical testing. The work focuses on identifying suitable sensor types for capturing PC failure signatures, and determining their optimal placement using DL techniques. Accordingly, the core challenges addressed include the identification of critical PC failure modes relevant to the selected case study, the associated monitoring challenges, and the integration of appropriate sensing solutions to detect and mitigate such failures.

\subsubsection{Failure Mode Classification and Prioritization}
\label{subsubsec:failure_mode_classification}
To establish a data-driven basis for the monitoring strategy, historical failure records from a field troubleshooting database were systematically analyzed. Based on frequency of occurrence and repair time metrics, failure modes were categorized into three major groups:
\begin{itemize}
    \item Mechanical failures, such as ceramic plate cracking, screw loosening and warpage in ceramic plates;
    \item Electrical failures, including open/short circuits and unstable contact resistance;
    \item Needle and pin failures, related to breakage or thermal damage.
\end{itemize}

\begin{figure}[htbp]
    \centering
    \includegraphics[width=1.0\columnwidth]{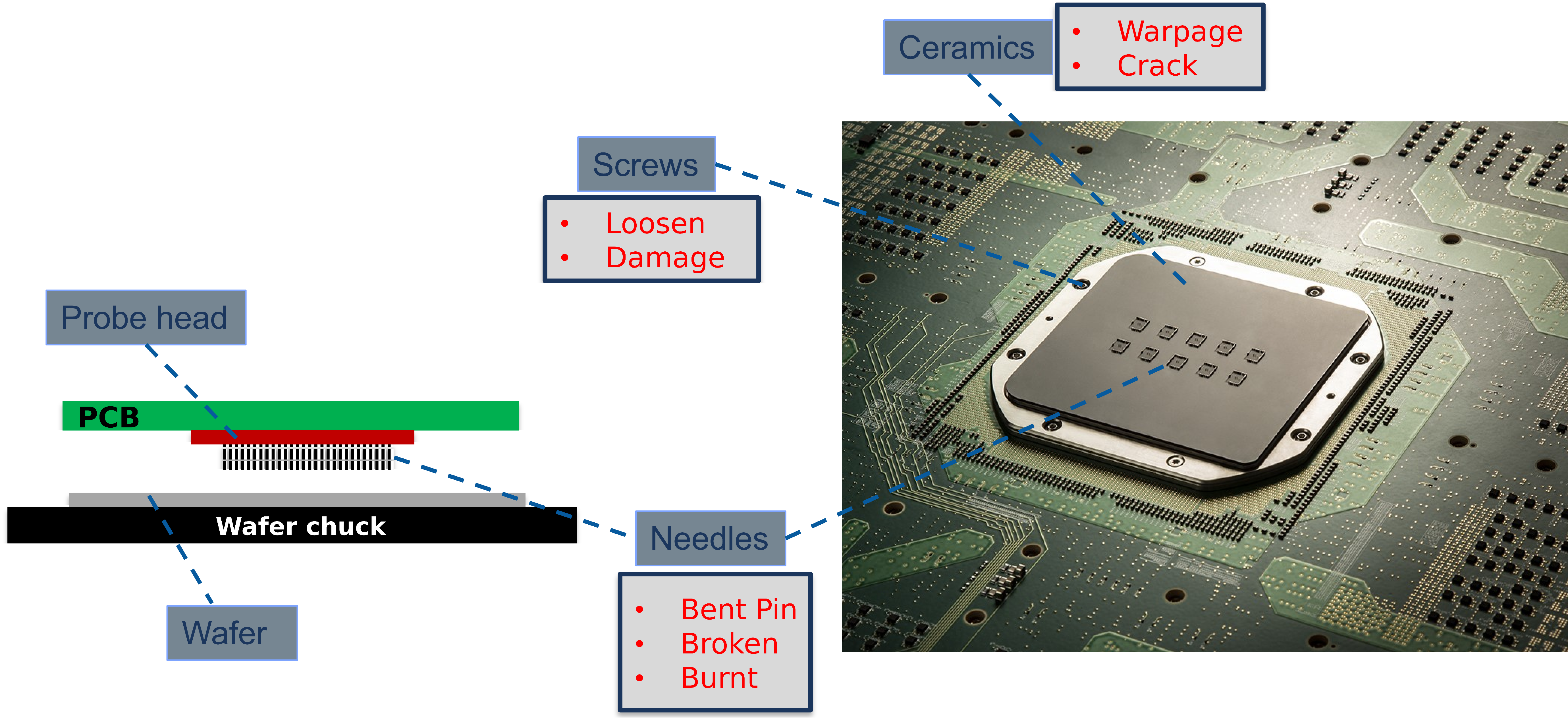}
    \caption{Probe Card Structure and Associated Failure Modes.}
    \label{fig:probe_card}
\end{figure}

Figure \ref{fig:probe_card} illustrates the structure of a PC and highlights the components where failures can occur. The diagram on the left displays the primary components: the probe head (PH), the Printed Circuit Board (PCB), and the needles, which are positioned above a wafer on a wafer chuck. The PH assembly comprises a stack of guide plates and a central housing, which together form the core structure of the assembly. This setup is used for testing wafers, by raising the wafer chuck to create a connection between the needles, inserted in the PH, and the pads on the wafer. The image on the right provides instead a detailed view of a PC, pointing out specific failure modes; these include warpage and cracks in the ceramic components, loosening or damage of the screws, and issues with the needles, such as bending-buckling, cracking, or burning.

\begin{figure}[htbp]
    \centering
    \includegraphics[width=0.9\columnwidth]{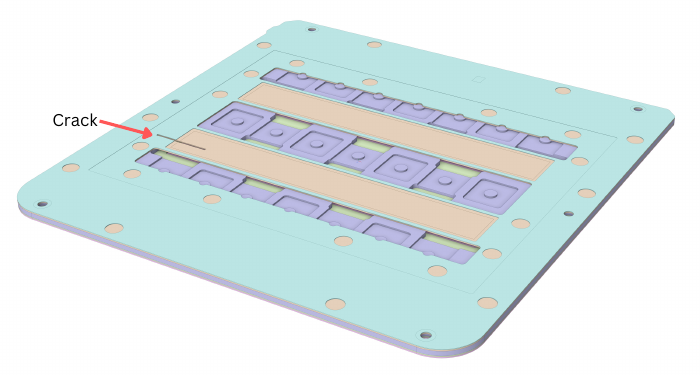}
    \caption{PH model in ANSYS, with a crack in the plate.}
    \label{fig:Model}
\end{figure}

\begin{figure}[htbp]
    \centering
     \includegraphics[width=0.8\columnwidth]{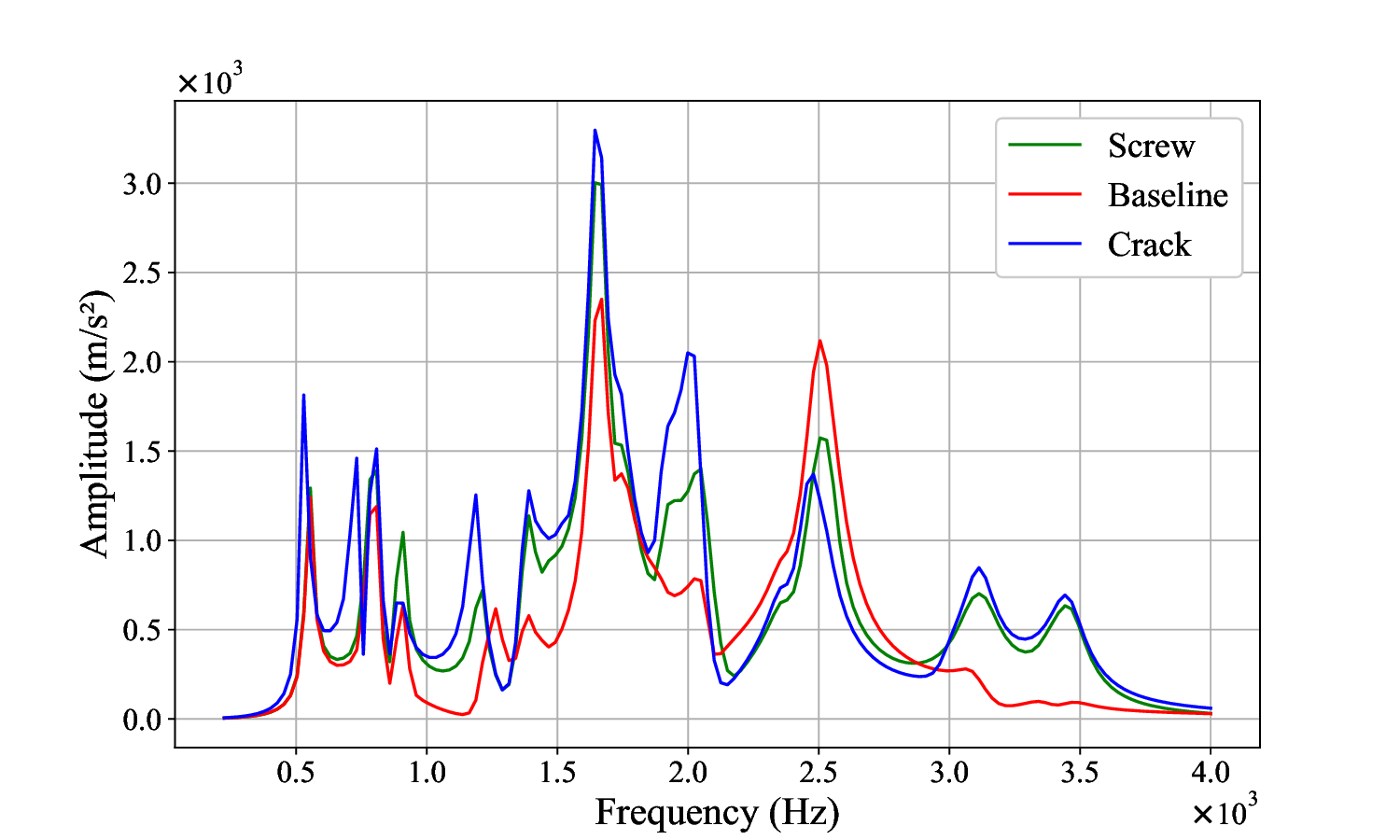}
    \caption{Comparison of PH FRF amplitudes for three different scenarios.}
    \label{fig:FRF}
\end{figure}

Based on the failure analysis and investigation into their root causes, the following major failure challenges were identified, along with proposed sensor-based mitigation strategies:
Based on the failure analysis and investigation into their root causes, the following major failure challenges were identified, along with proposed sensor-based mitigation strategies:

\begin{enumerate}
    \item[a)] \textbf{Needle and Card Contamination:} Contaminant buildup on needle tips can significantly compromise contact resistance and signal fidelity, ultimately reducing wafer test yield. Sources of contamination include residual aluminum oxide from wafer pads, previous test residues, and ambient particulates. To address this issues, accelerometers are deployed to monitor the cumulative number of touchdowns, enabling predictive and adaptive cleaning strategies and minimizing unplanned downtime through optimized maintenance scheduling.
    
    \item[b)] \textbf{Temperature Gradient-Induced Misalignments:} Temperature differences between PC and wafer surface are a leading cause of contact misalignment, contributing to both open and short circuit failures. These effects are exacerbated by insufficient soak times or improper thermal stabilization during the tests. Temperature sensors, strategically distributed across the PC, facilitate real-time thermal profiling and ensure early detection of abnormal gradients that could compromise test accuracy.
    
    \item[c)] \textbf{Cracks in Probe Card Plates:} Cracks in the PC structural plates pose a critical risk to mechanical integrity and probe tip alignment. Such cracks often nucleated near stress concentration points and can propagate undetected within the internal layers of the PH. Since only the lower plate is visible during operation, internal cracks require indirect detection. FE analyses using ANSYS Mechanical R2 2024 \citep{ref26} were conducted to simulate crack scenarios. Figure \ref{fig:Model} shows the FE model of the PH with an upper plate crack. Modal and harmonic response analyses revealed that cracks can significantly alter the FRF, particularly within the 400-4000 Hz band (audible spectrum), as illustrated in Figure \ref{fig:FRF}, where the spectra related to undamaged and damaged conditions are depicted. The outcome supports the use of microphones to capture acoustic signatures of the cracks. Recorded FRFs, discretized into 150 frequency points, are adopted as inputs for crack detection.
    
    \item[d)] \textbf{Loosening of Screws in the Probe Head Assembly:} The PH mechanical integrity is maintained by screws. Variations in screw torque, due for instance to vibrations generated during testing, could lead to tip misalignment and degraded contact. Loosened screws also elevate local stresses, increasing crack likelihood. Similar to cracks, screw loosening affects the mechanical coupling among the different structural components, causing measurable shifts in natural frequencies and FRFs. FE simulations confirmed that minor torque deviations produce distinguishable spectral changes, as shown in Figure \ref{fig:FRF}. This reinforces the use of vibration-based sensing microphones.
\end{enumerate}

\subsubsection{Proposed SHM Solution for Mechanical Failures}
\label{subsubsec:proposed_shm_solution}
\begin{figure}[htbp]
    \centering
    \includegraphics[width=0.8\columnwidth]{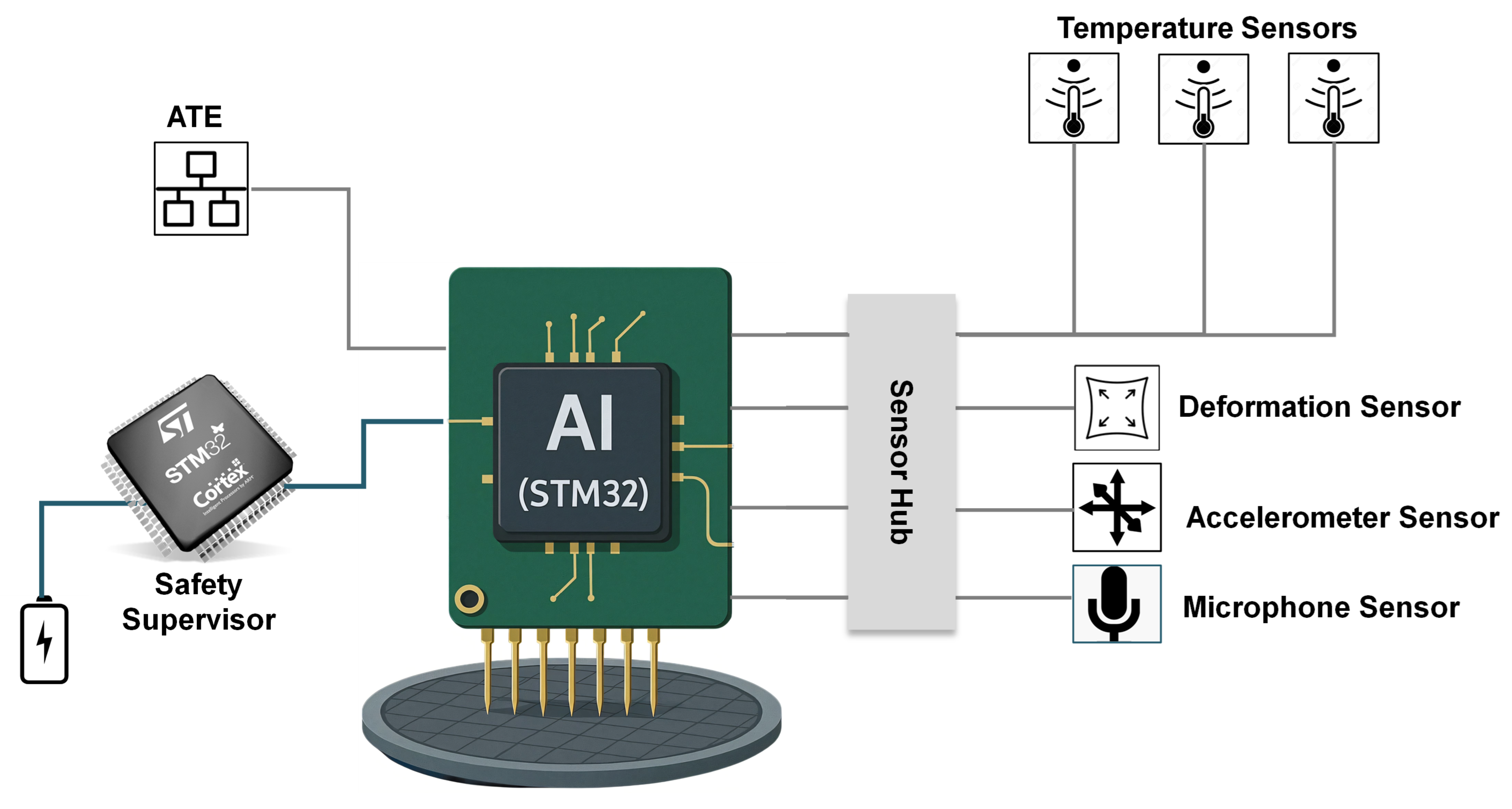}
    \caption{Sketch of the proposed sensor network.}
    \label{fig:Sen_net}
\end{figure}

By prioritizing the most critical failure modes and operational challenges of the PCs, this study identifies a tailored set of sensors to enable real-time SHM and predictive failure detection. Specifically, the following sensors  are proposed for integration:
\begin{itemize}
    \item \textbf{Accelerometers:} for counting probe touchdowns to estimate contact fatigue and optimize cleaning intervals, thereby supporting predictive maintenance strategies.
    \item \textbf{Microphones:} For capturing acoustic/vibrational signatures indicative of structural anomalies such as cracks or screw loosening, making them highly suitable for FRF-based SHM applications.
    \item \textbf{Temperature sensors:} for tracking thermal gradients across the PC surface and detecting misalignments arising from differential thermal expansion between the wafer and PC components.
\end{itemize}
While accelerometers and temperature sensors are part of a broader SHM vision, the current OSP work is specifically focuseed on FRF data from vibration sensors to detect mechanical failures. To implement online SHM and support predictive diagnostics, we developed a sensor network architecture comprising the above sensors, a primary microcontroller unit (MCU) responsible for data acquisition and the execution of embedded machine learning algorithms, and potentially an auxiliary ultra-low-power MCU for continuous system supervision. A conceptual illustration of this architecture is provided in Figure \ref{fig:Sen_net}. The continuously acquired data would serve for real-time anomaly detection and long-term failure forecasting, minimizing downtime and maintenance costs. In the initial deployment phase, sensor-embedded PCs are operated under real test conditions to acquire field data for training and validating the machine learning models. A significant design challenge is a robust detection of these mechanical damage modes: to this aim, we employ a digital shadow approach via FE modeling of the PH, to simulate degradations due to cracks and screw loosening. The corresponding FRFs recorded at candidate sensor locations are then exploited for OSP through DL and attention mechanisms.

\subsection{Dataset Generation and Preprocessing for OSP}
\label{subsec:dataset_generation}
The main objective for OSP is to determine the optimal number and placement of vibration-based sensors for monitoring the mechanical failures in PCs.
\subsubsection{Digital Shadow and FE Simulations for FRF Data Generation}
\label{subsubsec:digital_shadow}
Our data-driven DL approach required a comprehensive dataset. We developed a digital shadow specific for the studied PC, using ANSYS Mechanical R2 2024 FE models \citep{ref26}. Various failure scenarios related to cracks and screw-loosening were simulated, to compare the relevant responses against those of the undamaged baseline. Modal and harmonic response analyses were adopted to extract the FRFs, revealing that both failure modes significantly alter the vibrational behavior of the entire PC. Virtual sensors collected these FRFs for anomaly detection and classification of the failure modes. Additional details were already reported in \cite{ref53}.

\subsubsection{Candidate Sensor Locations}
\label{subsubsec:candidate_sensor_locations}
\begin{figure}[htbp]
    \centering
    \includegraphics[width=\columnwidth]{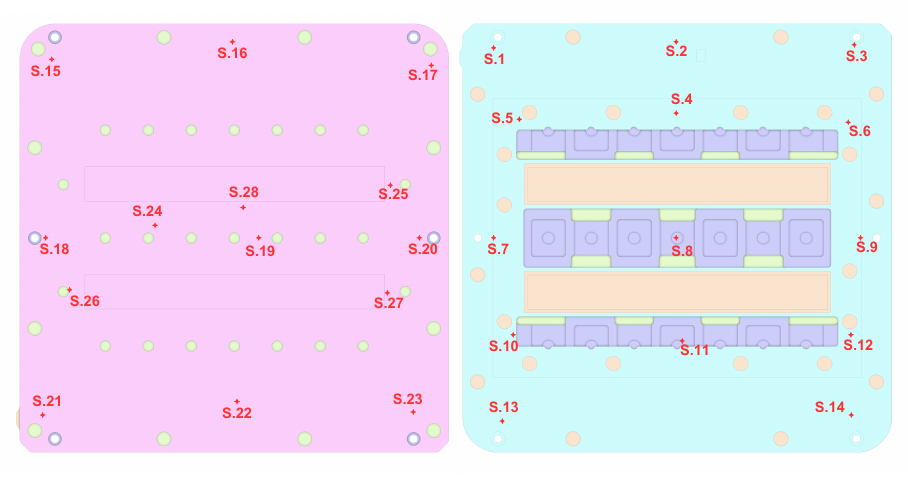}
    \caption{Representation of the 28 locations on the lower side (left) and upper side (right) of the PH, where FRFs were extracted.}
    \label{fig:sensors}
\end{figure}

For OSP, 28 potential sensor locations were identified in the FE model, see Figure \ref{fig:sensors}. FRF data were collected virtually at each sensor location, under the various scenarios introduced in what precedes and detailed in the following. As already anticipated, the final goal of this activity was to identify the most informative locations for detecting cracks and screw loosening using an attention-based DL model to classify damage and extract sensor attention scores.

\begin{figure}[htbp]
    \centering
    \includegraphics[width=\columnwidth]{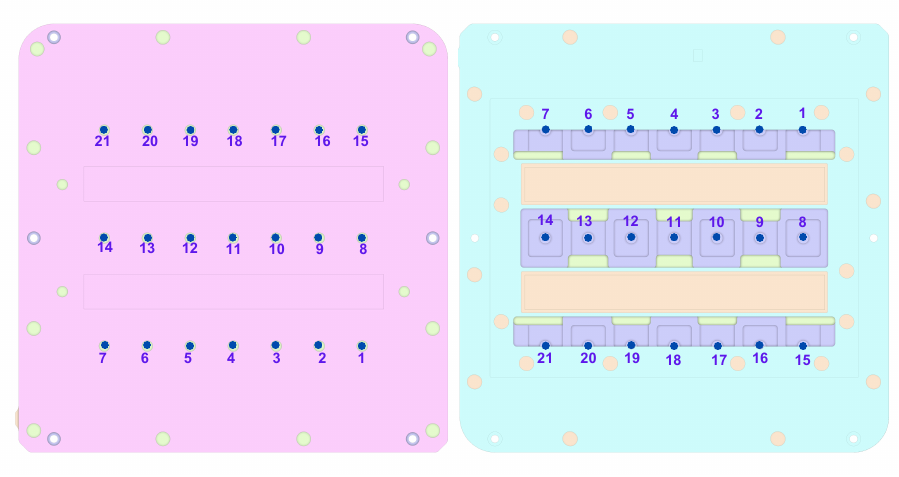}
    \caption{Adopted coding for the 21 screws used to connect the lower frame (left) to the housing (right) of the PH.}
    \label{fig:screws}
\end{figure}

\subsubsection{Data Challenges and Physics-Informed Scenario Expansion}
\label{subsubsec:data_challenges}
A significant challenge in the initial simulated dataset was the substantial imbalance between undamaged and damaged solutions. In fact, the simulations included: 21 scenarios representing loose screws as shown in the Fig. \ref{fig:screws}, with one screw loosened in each scenario; 8 crack scenarios, 4 in the upper plate and 4 in the lower plate of the PH; and only one single undamaged baseline case. This imbalance posed serious challenges for model training and generalization.

To address this imbalance and provide more training data for the DL models, we employed physics-informed scenario expansion techniques that systematically varied key physical parameters within realistic bounds. These expansions were based on fundamental physical principles rather than arbitrary mathematical transformations, ensuring that each new scenario actually represents a physically plausible condition. The scenario expansion incorporated three key physical variations, according to the following:
\begin{enumerate}
     \item[a)] \textbf{Material Property Variations:}  PHs exhibit a multi-component architecture, comprising heterogeneous materials with distinct mechanical properties. The guide plates, which contain precision-machined holes for probe positioning, are fabricated from Silicon Nitride (SiN) ceramic material. These guide plates are integrated with a housing structure that ensures mechanical stability and structural integrity, manufactured from Metal Alloy (MeAl). A controlled variation on Young's modulus and density of the SiN and MeAl components has been applied, to simulate a range of material conditions. Table \ref{tab:material_properties} summarizes these specific property variations. Although a statistical (Monte Carlo-like) procedure can account more properly of property variations, five distinct cases were selected in this work: one nominal case; one case with maximum property values; one case with minimum values; one case with minimum values for SiN and maximum for MeAl; and one case  with maximum values for SiN and minimum for MeAl. The five cases are all detailed in Table \ref{tab:2}.

    The property variations selected for SiN and the MeAl are rooted in the physical conditions of the manufacturing and fabrication processes for these material classes. For SiN, a greater variability is anticipated being a ceramic material. Its density can change due to differing levels of inherent porosity in the manufacturing process, justifying the ±5\% range. The Young's modulus of SiN is even more sensitive to its fabrication method, final crystalline structure, and the presence of impurities, making the ±10\% variation a reasonable estimate to encompass these effects. In contrast, the MeAl is assigned more constrained variations. Metal alloys typically exhibit a more consistent density related to the manufacturing process, hence the ±2\% variation looks reasonable. While its Young's modulus is influenced by the processing and the microstructure, the effect is generally less pronounced than in ceramics, making a ±5\% variation appropriate. By incorporating these physically-justified variations, we ensured that our simulations account for a realistic spectrum of material conditions, thereby enhancing the robustness and real-world applicability of our findings.
     \item[b)] \textbf{Environmental Conditions:} Five different temperatures of the environment ($25\,^{\circ}\mathrm{C}$, $100\,^{\circ}\mathrm{C}$, $150\,^{\circ}\mathrm{C}$, $200\,^{\circ}\mathrm{C}$, and $250\,^{\circ}\mathrm{C}$) were incorporated to account for thermal effects on the structural response.
     \item[c)] \textbf{Loading Conditions:} The mechanical load applied to the PH assembly represents the forces generated by the wafer overtravel during testing operations. To ensure proper electrical contact between the probe needles and the wafer pads, the chuck system establishes initial contact between the probes and the wafer surface, followed by an additional vertical displacement termed overtravel. During this process, each probe functions as a mechanical spring element, exerting controlled force on the wafer surface to break potential not-conductive layers. The total force must be precisely controlled to prevent wafer damage and ensure measurement accuracy. Five magnitudes of distributed load (0.1 MPa, 0.5 MPa, 1 MPa, 1.5 MPa, and 2 MPa) were simulated to represent different operational scenarios applied to the probe holder structure. These load levels correspond to varying chuck pressure requirements, with lower pressures (0.1-0.5 MPa) representing light contact conditions and higher pressures (1.0-2.0 MPa) simulating aggressive overtravel conditions for challenging contact scenarios. The distributed nature of the applied load represents a computational simplification of the realistic loading scenario where the chuck pressure is transmitted through the wafer surface and the probes to the PH assembly. In the FE simulations of the PH, this pressure is applied as a distributed surface load on the lower plates of the PH assembly for computational efficiency. This loading methodology enables the assessment of PH structural integrity and dynamic response characteristics under various operational conditions encountered in semiconductor testing environments.
\end{enumerate}
Combining all these cases, 125 variant conditions were generated for each initial scenario. This physics-informed expansion thus increased the dataset size while maintaining realism, to allow for 125 baseline cases, 1000 crack cases, and 2625 loose screw cases.

\begin{table}[htbp]
  \centering
  \caption{Adopted material property variations for SiN and MeAl.}
  \label{tab:material_properties}
  \small 
  \begin{tabular}{p{3.5cm}p{2.8cm}p{3.2cm}p{4.5cm}}
    \toprule
    \textbf{Property} & \textbf{Reference Value} & \textbf{Variation Range} & \textbf{Physical Justification for Variations} \\
    \midrule
    SiN Density (kg/m$^3$) & 3230 & $\pm 5\%$ (3069--3392) & effects of the manufacturing processes (in terms of, e.g. porosity). \\
    \midrule
    SiN Young's Modulus (GPa) & $310 $ & $\pm 10\%$ ($279 $--$341 $) & effects of fabrication, crystalline structure, impurities. \\
    \midrule
    MeAl Density (kg/m$^3$) & 8150 & $\pm 2\%$ (7987--8313) & effects of manufacturing processes. \\
    \midrule
    MeAl Young's Modulus (GPa) & $145$ & $\pm 5\%$ ($138 $--$152 $) & effects of processing, and microstructure. \\
    \bottomrule
  \end{tabular}
\end{table}

\begin{table}[htbp]
  \centering
  \caption{Material variations-induced cases.}
  \label{tab:2}
  \small 
  \begin{tabular}{cp{2.2cm}p{2.8cm}p{2.2cm}p{2.8cm}}
    \toprule
    \textbf{Case} & \textbf{SiN Density (kg/m$^3$)} & \textbf{SiN Young's Modulus (GPa)} & \textbf{MeAl Density (kg/m$^3$)} & \textbf{MeAl Young's Modulus (GPa)} \\
    \midrule
    1 & 3230 & $310 $ & 8150 & $145 $ \\
    2 & 3069 & $279 $ & 7987 & $138 $ \\
    3 & 3392 & $341 $ & 8313 & $152 $ \\
    4 & 3069 & $341 $ & 8313 & $138 $ \\
    5 & 3392 & $279 $ & 7987 & $152 $ \\
    \bottomrule
  \end{tabular}
\end{table}

\subsubsection{Statistical Balancing and Physics-Aware Data Augmentation}
\label{subsubsec:statistical_balancing}
Despite the physics-informed expansion of the scenarios, a class imbalance remained in the dataset. To address this during training, we implemented a Synthetic Minority Over-sampling Technique (SMOTE). SMOTE generates synthetic examples for minority classes by interpolating between existing samples, thereby balancing the class representation during training. In addition to SMOTE, applied only to the training data within each cross-validation fold, we employed a physics-aware statistical data augmentation during training to enrich the dataset. This process ensures that the model learns from a wider, more realistic range of data, improving its robustness and generalization. The parameters for this augmentation were derived directly from the statistical properties of the original dataset to ensure they are physically meaningful. Three augmentation techniques were applied to the training data in each fold:

\begin{enumerate}
    \item[a)] \textbf{Gaussian Noise Addition:} To simulate plausible measurement noise and inherent sensor variability, random perturbations following a normal distribution were added to each amplitude of the FRF. The standard deviation ($\sigma$) of this noise was not chosen arbitrarily but was computed as 10\% of the standard deviation of the amplitude data for each specific class, ensuring the parameters are physically meaningful. This resulted in $\sigma = 119.86~\text{m/s}^2$ for the baseline, $\sigma = 119.37~\text{m/s}^2$ for screw defects, and $\sigma = 119.27~\text{m/s}^2$ for crack defects. This level of noise is minor compared to the overall signal amplitude---for instance, the primary resonance peaks in the figure reach amplitudes of over $2000~\text{m/s}^2$. The applied noise, therefore, realistically simulates sensor interference without obscuring the underlying structural features of the FRF that are critical for classification.
    \item[b)] \textbf{Frequency Jitter:} To account for slight variations in sensor calibration or measurement timing, controlled shifts were introduced by rolling the signal along its frequency axis. A maximum shift of 3 data points was used, corresponding to 2\% of the 150 frequency steps in our data. This technique simulates small shifts in the observed resonant peaks along the frequency axis, a common real-world phenomenon, without altering the shape of the response curve.
    \item[c)] \textbf{Amplitude Scaling:} To reflect variations in excitation force or sensor sensitivity during testing, signal amplitudes were uniformly scaled by a random factor within a ±10\% range for all the classes. As seen in Figure \ref{fig:FRF}, different structural conditions naturally lead to variations in the overall FRF amplitude: this scaling range thus introduces plausible amplitude variations consistent with physical measurements, while preserving the fundamental shape and relative peak heights that characterize each specific condition.
\end{enumerate}
This combination of one-time physics-informed scenario expansion, SMOTE, and dynamic physics-aware statistical data augmentation on training folds helped enhance the representational capacity and class balance.

\subsubsection{Final Dataset Characteristics}
\label{subsubsec:final_dataset_characteristics}
Compared to our prior research \citep{ref53}, the current dataset exhibits substantial expansion across several key parameters: material cases increased from 3 to 5, temperature conditions from 3 to 5, and loading conditions from 3 to 5. Consequently, instead of the 810 samples adopted in our previous work, the current study utilizes 3750 samples. The combined dataset thus consists of 3750 samples, each one collecting the measurements from the 28 sensors, with each sensor providing data at 150 frequency values. This diverse dataset provides a stronger foundation for training deep learning models, intended to be robust and exhibit good generalization capabilities to unseen data. It is well understood that the importance of a large and diverse dataset in DL cannot be overstated, as it allows the model to learn more generalizable representations and reduces the risk of overfitting to specific characteristics of a limited dataset.

\subsection{Model Architecture: Hybrid CNN-Transformer  (TransformerSHM)}
\label{subsec:model_architecture}
\begin{figure}[htbp]
    \centering
    \includegraphics[width=0.6\columnwidth]{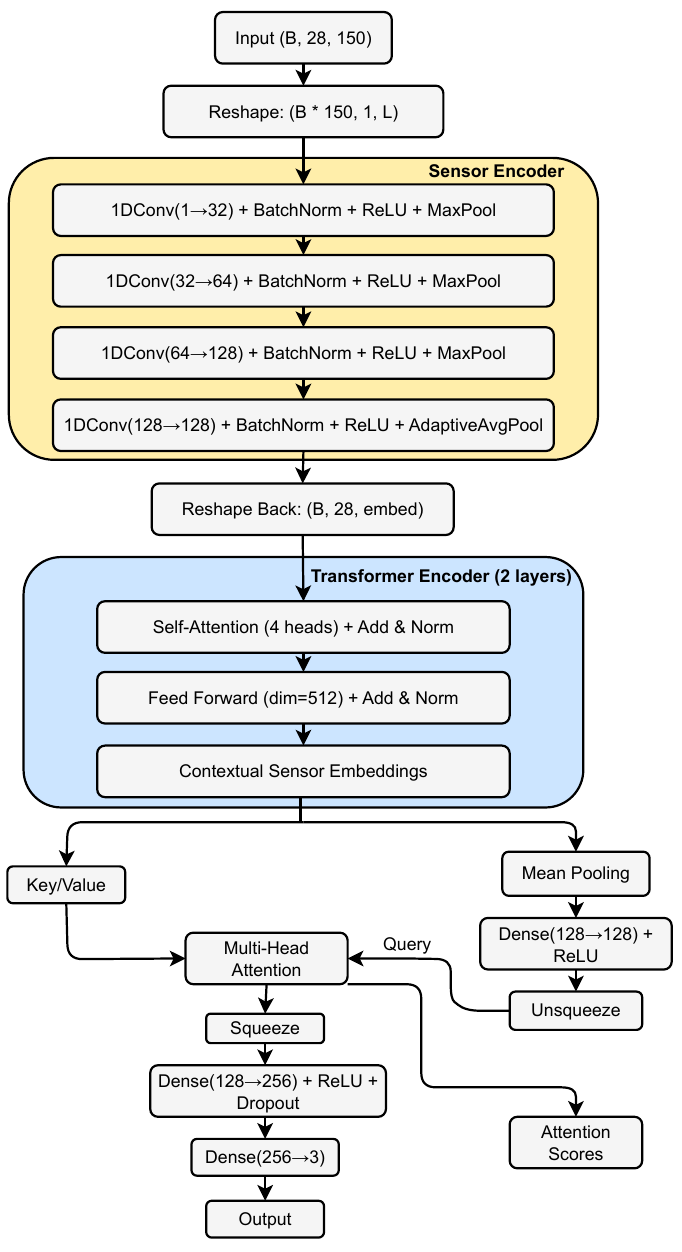}
    \caption{Scheme of the DL architecture with the attention mechanism.}
    \label{fig:model_arch}
\end{figure}

This study implements a novel hybrid deep learning architecture that synergistically combines one-dimensional convolutional neural networks (1D CNNs) with Transformer-based sequence modeling, designated as TransformerSHM. The proposed architecture addresses the inherent limitations of traditional approaches by leveraging the local feature extraction capabilities of CNNs and the global dependency modeling strengths of Transformers for structural health monitoring applications. The model comprises three principal components: a sensor encoder module, a Transformer-based integration module, and an attention-based classification module, as illustrated in  Figure \ref{fig:model_arch}.

\subsubsection{Sensor Encoder Module}
The sensor encoder module processes individual sensor channels containing FRF data independently through a cascaded 1D convolutional architecture. Each sensor input, characterized by dimensions (B, 1, 150) where B represents the batch size and 150 denotes the sequence length, undergoes feature extraction through four sequential 1D convolutional layers. The convolutional layers employ a kernel size of 3 with padding of 1 to preserve spatial dimensions, while the output channels progressively increase from 32 to 64, 128, and 128 channels, respectively. This hierarchical feature extraction strategy enables the capture of increasingly complex and abstract representations at each level. Each convolutional operation is succeeded by batch normalization and ReLU activation functions, followed by max pooling operations with a kernel size of 2 for dimensionality reduction. The batch normalization technique serves dual purposes: stabilizing the training process by normalizing layer inputs and accelerating convergence by reducing internal covariate shift. The ReLU activation function introduces essential non-linearity, enabling the network to learn complex mappings between input features and target outputs. Max pooling operations enhance model robustness against minor signal variations while reducing computational complexity. The rationale for employing 1D CNNs in the sensor encoder lies in their inherent capability to capture local temporal patterns and dependencies within the 150-point frequency response sequences. Unlike fully connected layers that treat each input dimension independently, convolutional operations preserve spatial relationships and enable the detection of localized features characteristic of structural anomalies. The final adaptive average pooling layer reduces the variable-length sequences to fixed-length embedding vectors of dimension 128, ensuring consistent representation regardless of input sequence variations.

\subsubsection{Transformer-based Integration Module}
The Transformer-based integration module processes the sequence of sensor embedding vectors with input dimensions (B, 28, 128), where 28 corresponds to the total number of sensors deployed in the monitoring system. The module implements a multi-layer Transformer encoder architecture consisting of two encoder layers, each incorporating multi-head self-attention mechanisms with four attention heads. The integration of positional encodings with sensor embeddings provides spatial awareness regarding sensor positioning and arrangement within the structural system. Each Transformer encoder layer comprises two primary sub-components: a multi-head self-attention mechanism and a position-wise feed-forward network, both incorporating residual connections and layer normalization. The selection of Transformer architecture for sensor integration addresses fundamental limitations of sequential processing approaches. Unlike recurrent neural networks that process information sequentially, the self-attention mechanism enables simultaneous modeling of relationships across the entire sensor array. This parallel processing capability is particularly advantageous for SHM applications, where damage manifestations may involve complex spatial correlations between non-adjacent sensors. The multi-head self-attention mechanism allows each sensor's embedding to dynamically attend to all other sensor embeddings, computing attention weights that reflect the relevance of inter-sensor relationships for damage detection. This approach is essential for identifying distributed damage patterns that may not be localized to specific sensor locations but rather manifest through subtle correlations across the sensor network.

\subsubsection{Attention-based Classification Module}
The classification module processes the contextualized sensor embeddings generated by the Transformer encoder through an additional attention mechanism designed to identify the most informative sensors for damage classification. A query vector is derived from the mean of the Transformer output sequence and subsequently processed through a dense layer with ReLU activation to enhance representational capacity. This query vector engages in multi-head attention with the Transformer encodings using four attention heads, generating sensor-specific attention weights that quantify the relative importance of each sensor for the classification task. The attention weights facilitate a weighted aggregation of sensor encodings, creating a comprehensive representation that emphasizes the most discriminative sensor information. The aggregated representation is processed through a final classification network comprising a fully connected layer with ReLU activation, dropout regularization for overfitting prevention, and a terminal dense layer for multi-class prediction of structural condition states. The attention-based classification approach provides dual benefits for SHM applications. Primarily, it enables effective aggregation of rich, contextualized sensor information into an optimal representation for classification. Additionally, the attention weights offer direct interpretability regarding sensor importance, facilitating identification of critical sensor locations and supporting OSP strategies. This interpretability transforms the model from a black-box predictor into a tool for actionable engineering insights, enabling practitioners to understand which sensors contribute most significantly to damage detection decisions.

\subsection{Cross-Validation Framework and Training Protocol}
\label{subsec:cross_validation}
A 10-fold stratified cross-validation framework ensured model reliability and generalization, maintaining consistent class distribution across folds, critical for our imbalanced dataset. For each fold, the data were partitioned into training/validation. SMOTE was applied only to the training portion, followed by physics-aware statistical augmentation. Custom augmentation factors were dynamically calculated based on the class distribution after SMOTE, aiming to further balance the data and diversify the training set. These custom augmentations included Gaussian noise, jitter, and scaling. The validation set remained untouched by SMOTE and these augmentations for unbiased evaluation. To ensure overall reliability and robust model evaluation, the K-fold cross-validation procedure was repeated three times, each with different initial data shuffles. Within each specific fold of these repetitions, an ensemble of ten models was trained, each model initialized with a different random seed. Training utilized early stopping (30-epoch patience on validation loss) to prevent overfitting, and learning rate reduction on plateau to enhance convergence.

Training used cross-entropy loss with class weights, and L1 regularization on final classification attention weights. This specific loss criterion was employed as it is suitable for multi-class classification problems, and internally applies a Softmax activation to the model's output (logits) before computing the negative log-likelihood loss. This integrated approach ensures numerical stability during training. 
The cross-entropy loss is defined as:
\begin{equation}
\label{eq:cross_entropy}
\mathcal{L} = - \frac{1}{N} \sum_{i=1}^{N} \sum_{c=1}^{C} y_{ic} \log(p_{ic})
\end{equation}
where: $N$ is the number of samples; $C$ is the number of classes; $y_{ic}$ is a binary indicator (0 or 1) if class label $c$ is the correct classification for the $i$-th observation; and  $p_{ic}$ is the predicted probability that $i$-th observation belongs to class $c$ (obtained after applying the Softmax function to the logits). Class weights, derived from the inverse frequency distribution of the respective training data (after SMOTE and custom augmentation) for each fold, were incorporated into this loss function to further address data imbalance, ensuring that minority classes contribute more significantly to the loss and receive greater attention during training.

Additionally, L1 regularization was applied to the attention weights from the Multihead Attention layer of the TransformerSHM model. This technique adds a penalty to the loss function proportional to the sum of the absolute values of the model weights. The L1 regularization term is defined as:
\begin{equation}
\label{eq:l1_penalty}
L1_{\text{penalty}} = \lambda \sum_{j=1}^{M} |w_j|
\end{equation}
where: $\lambda$ is the regularization coefficient, which controls the strength of the penalty (set to $1 \times 10^{-4}$ in our training); $w_j$ represents the individual attention weights; and $M$ is the total number of attention weights being regularized. The total loss function for training was therefore a combination of the weighted cross-entropy loss and the L1 penalty. By penalizing the absolute magnitude of the attention weights, L1 regularization encourages sparsity, meaning it drives many of the weights to become exactly zero. In the context of our model, this promotes sensor selectivity. The model is incentivized to rely more heavily on a few highly informative sensors rather than distributing importance across many, thereby enhancing the interpretability of sensor contributions to the final prediction.

This comprehensive approach, leveraging an ensemble of 10 models per fold within a triply repeated K-fold framework, was assumed to provide robust predictions and facilitate uncertainty estimation. Training used cross-entropy loss with class weights, L1 regularization on final classification attention weights (promoting sensor selectivity), and the Adam optimizer (learning rate of $1 \times 10^{-4}$). Performance assessment used accuracy, balanced accuracy, precision, recall, F1-score, and Receiver Operating Characteristic Area Under the Curve (ROC-AUC).

\subsection{Sensor Importance Analysis}
\label{subsec:sensor_importance_analysis_methodology} 
A distinctive feature of the proposed methodology is the incorporation of sensor importance analysis through the model attention mechanism. The multi-head attention weights generated during inference provide valuable insights into which sensors contribute most significantly to defect detection and classification. For each sample, attention weights across all heads were averaged to produce per-sensor importance scores, which were then aggregated across the validation set to determine the overall sensor importance. This analysis enabled several practical advantages: identification of critical sensor locations most sensitive to specific structural defects; interpretability of the model decisions; and assessment of consistency in sensor utilization across different data partitions, through standard deviation analysis of importance scores. Sensors with consistently high importance scores represented reliable indicators for SHM, while high variance may indicate context-dependent utility. The sensor importance analysis directly addressed our primary research objective of OSP, providing a data-driven approach to reducing the number of required sensors while maintaining high detection accuracy. This optimization has obvious implications for cost reduction, system simplification, and practical deployment of SHM systems in real-world PC applications.

\section{Results: Performance Evaluation and Sensor Importance Analysis}
\label{sec:results}
The experimental results reported next are going to highlight the efficacy of the proposed CNN-Transformer hybrid model, or TransformerSHM, in classifying PC states and elucidating sensor contributions. Leveraging the dataset made of 3,750 base samples and a rigorous evaluation framework consisting of 3 repetitions of a 10-fold stratified cross-validation, the model demonstrates robust performance. In this section, we analyze the relevant classification, robustness, learning dynamics, and sensor importance.

\subsection{Overall and Class-wise Classification Performance}
\label{subsec:classification_performance}
The overall performance of the TransformerSHM model, aggregated across all the 3 repetitions of 10-fold cross-validation, is presented in Table \ref{tab:performance_compact}. The model achieves an accuracy of 99.83\%, reflecting its ability to correctly classify PC states in the majority of the cases. To account for potential class imbalance, balanced accuracy is reported at 99.86\%, indicating consistent performance across all classes. The Receiver Operating Characteristic Area Under the Curve (ROC AUC) of nearly 1.00 (0.9999) further underscores the excellent discriminatory power of the model, demonstrating its capability to distinguish between different failure states. These metrics collectively suggest that the model is both accurate and reliable, making it a viable tool for failure detection in probe card monitoring. In particular, the high ROC AUC value indicates that the model maintains strong performance even when the classification threshold is varied, a critical attribute for applications where false negatives (missed failures) carry significant consequences.

\begin{table}[htbp]
  \centering
  \caption{Classification performance metrics (averaged over the 30 models).}
  \label{tab:performance_compact}
  \begin{tabular}{lccc}
    \hline
    \multicolumn{4}{c}{\textbf{Overall Performance}} \\
    \hline
    Accuracy (\%) & 99.83 & & \\
    Balanced Accuracy (\%) & 99.86 & & \\
    ROC-AUC & 0.9999 & & \\
    \hline
    \multicolumn{4}{c}{\textbf{Class-wise Performance}} \\
    \hline
    & Precision & Recall & F1-score \\
    \hline
    Baseline & 0.959 & 1.0 & 0.979 \\
    Loose Screw & 1.0 & 0.999 & 0.999 \\
    Crack & 0.999 & 0.997 & 0.998 \\
    \hline
  \end{tabular}
\end{table}

Class-specific performance metrics, detailed in Table \ref{tab:performance_compact}, provide deeper insights into the model behavior across the three classes: Baseline (healthy state), Loose Screw, and Crack. For the Baseline class, the model achieves a high precision of 0.959 (95.9\%). Remarkably, its recall is 1.0 (100\%), indicating that all the healthy cases are correctly identified, and none are misclassified as failure states. The F1-score of 0.979 reflects the excellent overall performance for this class. For the Loose Screw class, the model exhibits virtually perfect performance with a precision of 1.0 (100\%) and an outstanding recall of 0.999 (99.9\%). This translates to an F1-score of 0.999, signifying that the model is exceptionally able at identifying this common failure mode while essentially eliminating false positives. The robust and near-perfect detection of loose screws is particularly valuable, as this failure type can lead to misalignment and wafer damage if undetected. The Crack class, representing a critical failure mode, also shows exceptional performance. The model achieves a precision of 0.999 (99.9\%), meaning that false positives for cracks are extremely rare. Its recall of 0.997 (99.7\%) is again exceptionally high, ensuring that nearly all the crack instances are correctly identified. This high recall is crucial in semiconductor manufacturing, where missing a crack could result in significant downstream defects and economic losses. The F1-score of 0.998 reflects an outstanding balance, prioritizing the detection of these critical failures with high accuracy. These class-wise results further highlight the model ability to tailor its performance to the practical needs of PC monitoring, balancing reliability for common failures with sensitivity to rare but severe ones.

\begin{figure}[htbp]
    \centering
    \includegraphics[trim=0 0 3cm 0.9cm, clip,  width=0.5\columnwidth]{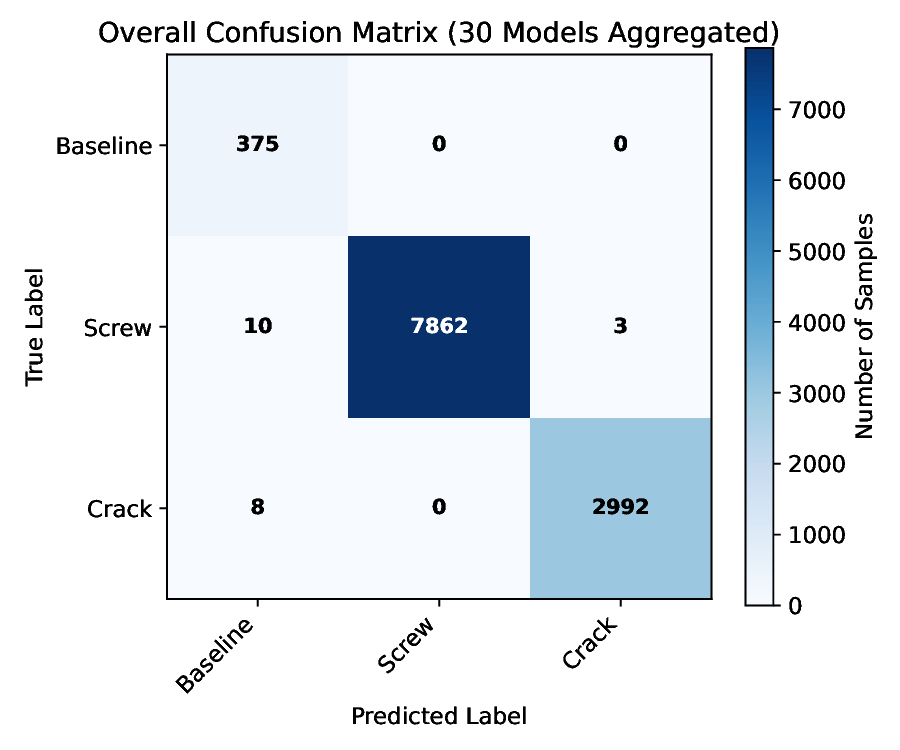}
    \caption{Overall confusion matrix, as aggregated from 3 repetitions of the 10-fold cross-validation.}
    \label{fig:conf_matrix}
\end{figure}

The aggregated confusion matrix (Figure \ref{fig:conf_matrix}), representing the combined predictions from all the validation folds across the 3 repetitions, totaling 11,250 evaluated samples, shows the detailed classification behavior: all the 375 baseline instances were correctly classified, demonstrating a perfect identification achievement; 7,864 out of 7,875 loose screw instances correctly identified, with only 8 instances misclassified as baseline and  3 as cracks; 2,992 out of 3,000 crack instances were correctly classified, with 8 instances misclassified as baseline and none as loose screws. This detailed analysis through the confusion matrix underscores the model outstanding classification accuracy across all the states. Misclassifications are exceedingly rare: the Baseline state is perfectly identified, and the distinctions between Loose Screw and Crack states are made with very high precision and recall, indicating minimal confusion. These results suggest that the features derived from the FRF patterns, as processed by the TransformerSHM model, are highly discriminative for the defined health states.

\subsection{Model Robustness and Learning Dynamics}
\label{subsec:model_robustness}
\begin{figure}[htbp]
    \centering
    \includegraphics[ width=0.7\columnwidth]{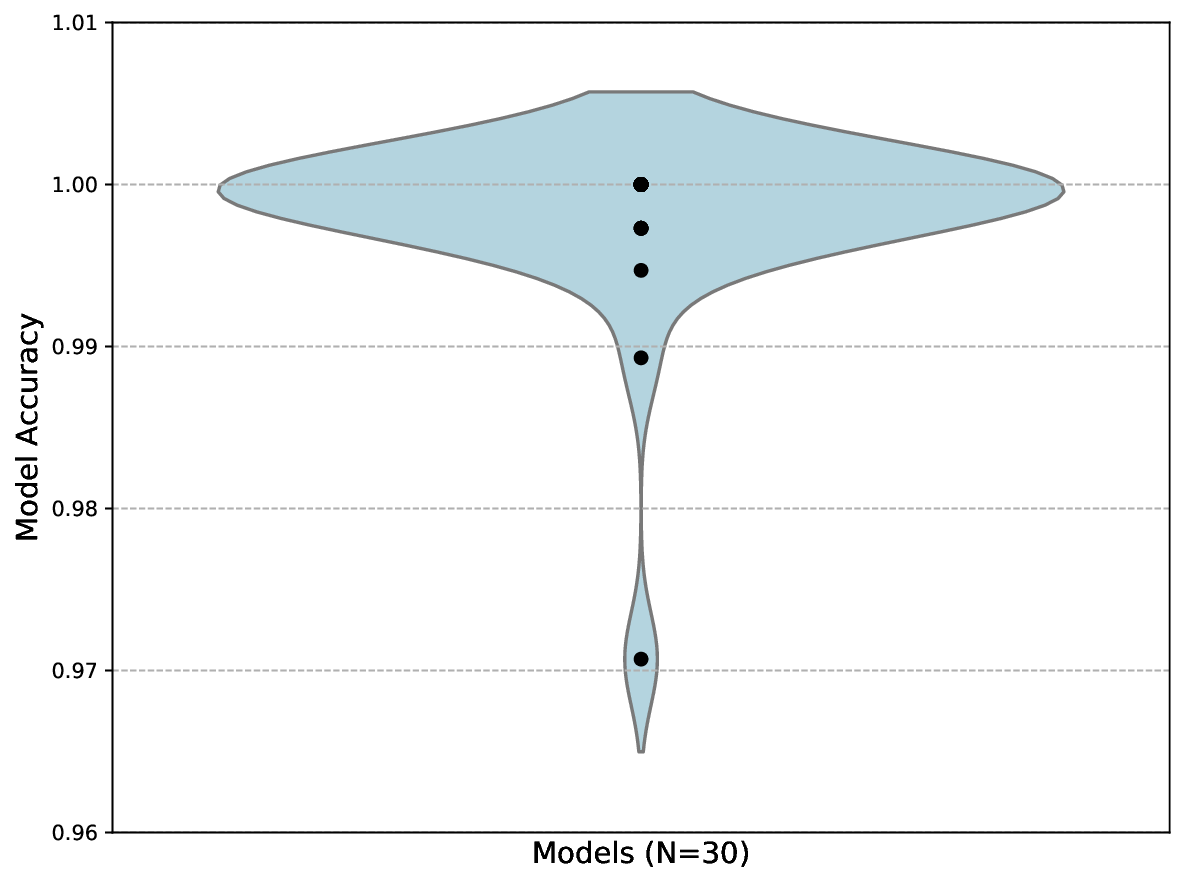}
    \caption{Accuracy distribution across the 30 models.}
    \label{Acc-Dist}
\end{figure}

To assess the model stability across different data partitions and training initializations, a boxplot of the accuracy scores for all the 30 trained models is presented in Figure \ref{Acc-Dist}. The tight distribution of the resulting accuracies, with minimal spread and no significant outliers, indicates that the model performs consistently, regardless of how the data is split or how the model training is initialized. This robustness is a direct result of the adopted stratified cross-validation approach, repeated across different data shuffles to ensure varied data partitions, combined with the use of multiple distinct random seeds for model training across repetitions, to mitigate variability from specific starting conditions. The consistent performance across all runs provides an idea regarding the confidence in the model generalizability, making it suitable for deployment in varied operational environments where PC conditions may differ.

\begin{figure}[htbp]
    \centering
    \includegraphics[trim=0 0 0 0.9cm, clip, width=1\columnwidth]{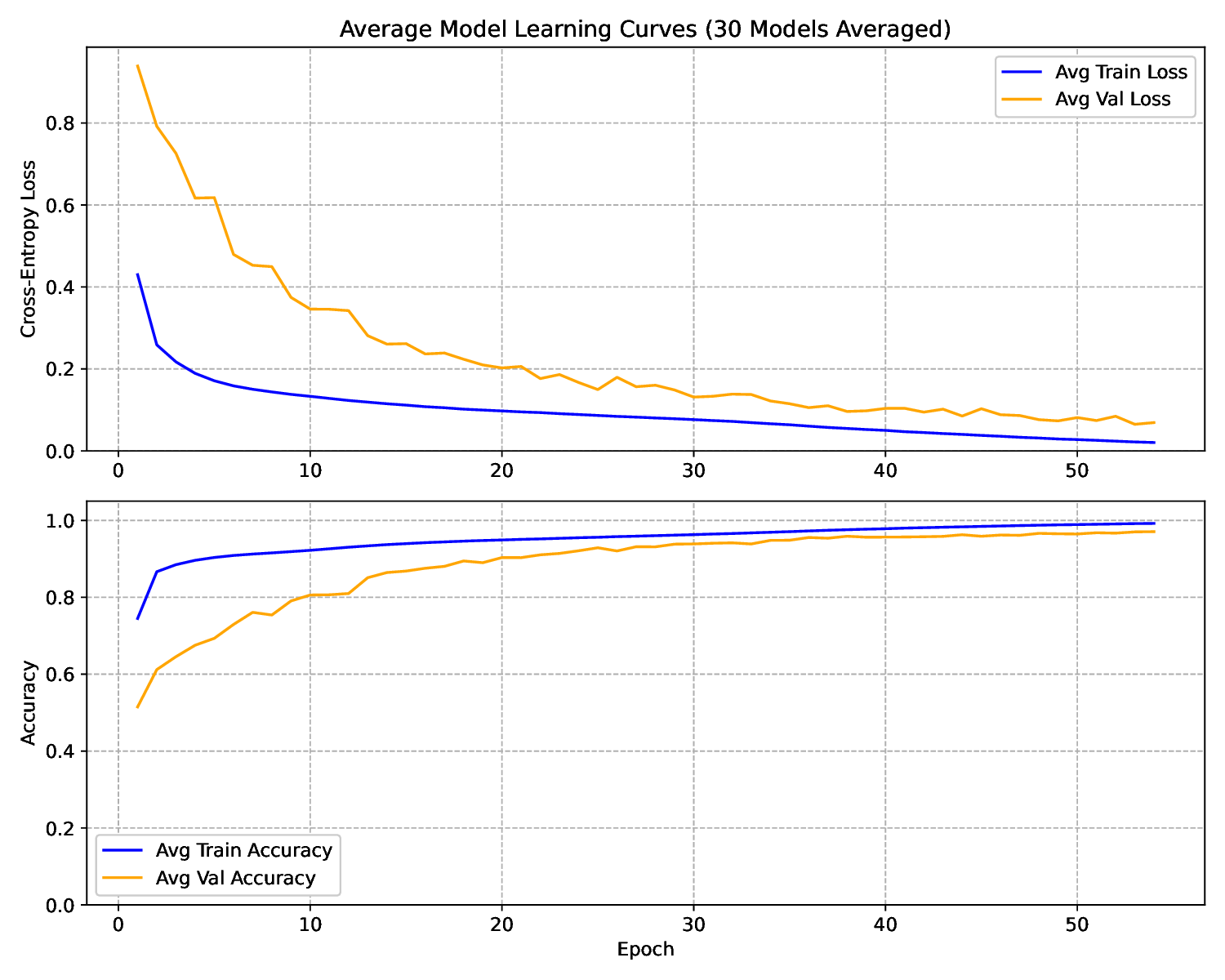}
    \caption{Model learning curves averaged over the 30 models.}
    \label{Lr-curve}
\end{figure}

The average learning curves, illustrated in Figure \ref{Lr-curve}, track the model training and validation loss and accuracy over epochs, again as averaged across all the 30 trained models. The curves show a steady decrease in the loss and a corresponding increase in the accuracy for both the training and validation sets, with both metrics converging effectively by the end of training. The minimal gap and absence of divergence between training and validation curves suggest that the model is not overfitting, a common concern with DL models, even on augmented datasets. This effective learning behavior is attributable to the careful application of data augmentation only to the training set within each fold, preserving the integrity of the validation process. Such a dynamics confirms that the TransformerSHM model learns meaningful patterns from the FRF data, leveraging the CNN local feature extraction and the Transformer global dependency modeling to achieve high performance without memorizing noise or artifacts.

\subsection{Sensor Importance Analysis}
\label{subsec:sensor_importance_analysis_results} 
\begin{figure}[htbp]
  \centering
   a{\includegraphics[trim=0 0 0 0.9cm, clip, width=1\linewidth]{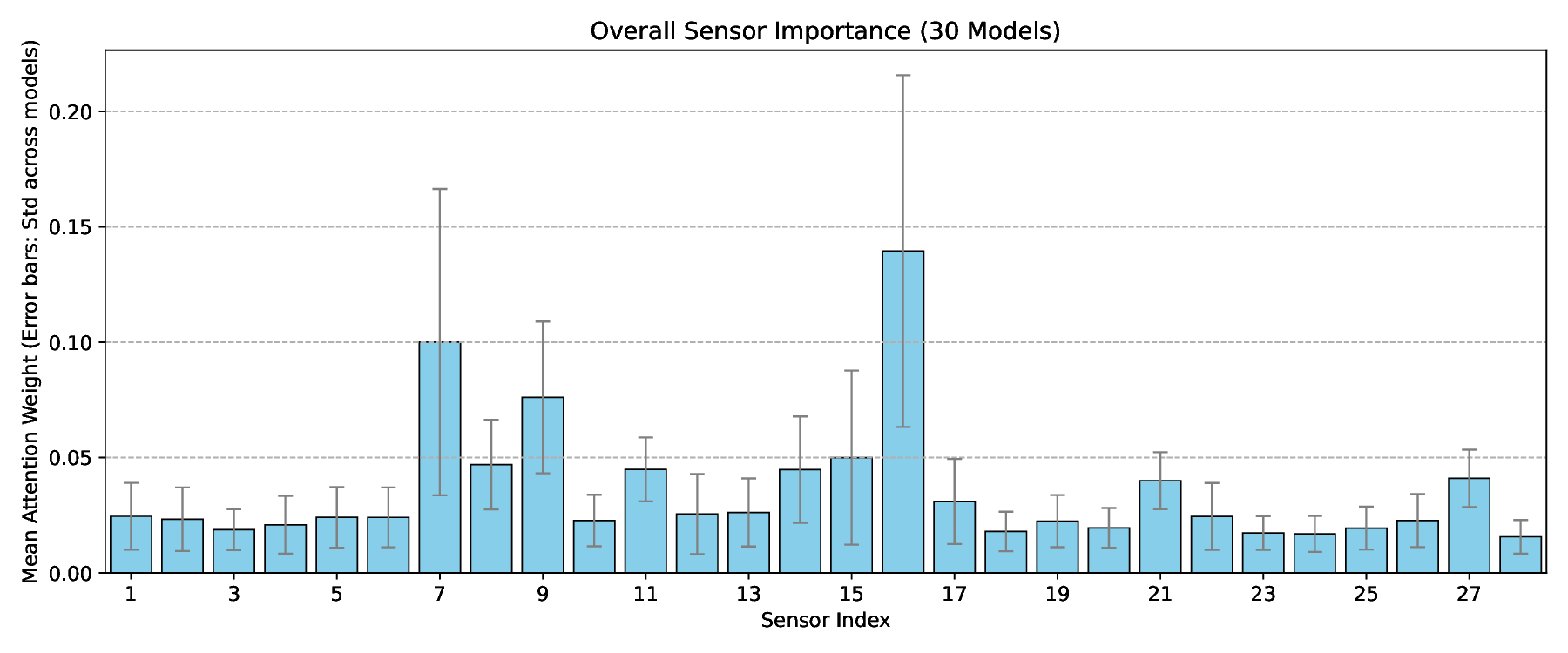}\label{fig:sensor_ranking}}\\[1ex]
   b{\includegraphics[trim=0 0 0 0.9cm, clip, width=1\linewidth]{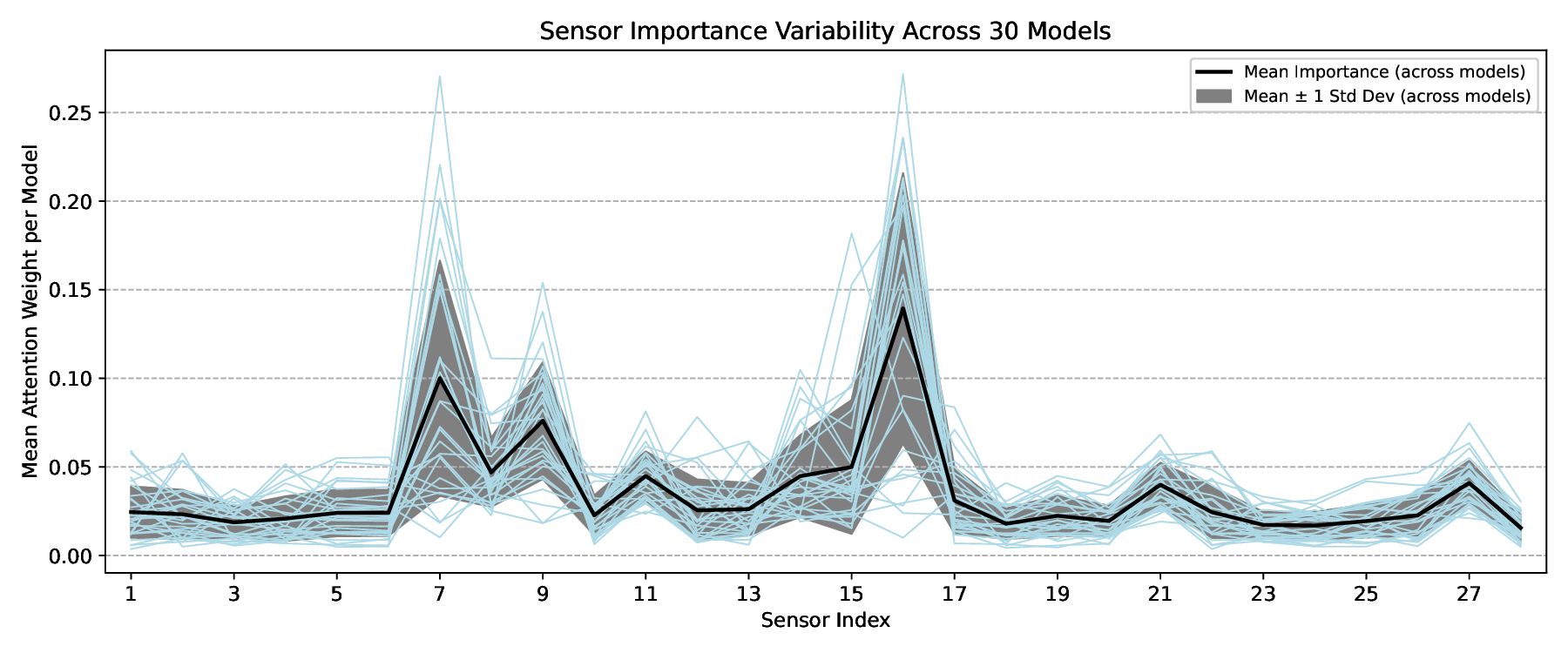}\label{fig:sensor_variability}}
  \caption{Sensor importance analysis: (a) overall sensor importance; (b) sensor importance variability.}
  \label{fig:sensor}
\end{figure}

A key advantage of the TransformerSHM model is its attention mechanism, which provides interpretable insights into sensor contributions. Figure \ref{fig:sensor}a displays the mean attention weight per sensor, averaged across all the validation samples from the 30 trained models, with error bars indicating the standard deviation of these scores. Sensor 16 consistently exhibits the highest attention weight, with sensors such as 7 and 9 also demonstrating significant contributions and low variability, suggesting they can play a pivotal role in failure detection. This finding is corroborated by Figure \ref{fig:sensor}b, which plots individual model results, showing stable importance rankings across the different training iterations.

The prominence of Sensor 16, along with sensors like 7 and 9, likely reflects their strategic placement within the probe card as shown in Figures \ref{fig:sensors} and \ref{fig:screws}, capturing critical vibrational signals indicative of failure. These sensors are placed near the ceramic plate interfaces or screw attachment points, where cracks and loosening are most likely to manifest. The low variability in their attention weights across the different model runs further validates their reliability, making them prime candidates for inclusion in an optimized sensor network. These insights have practical implications for sensor network design: by prioritizing high-importance sensors, manufacturers can potentially reduce the number of sensors deployed without sacrificing detection accuracy, lowering costs and simplifying maintenance workflows. This interpretability sets the TransformerSHM model apart from traditional CNN-based approaches, offering actionable guidance beyond mere classification. The reported findings not only validate the methodological advancements introduced in this study, but also pave the way for a practical implementation in semiconductor manufacturing, enhancing yield and reliability.

\section{Discussion}
\label{sec:discussion}
This study has addressed PC SHM and failure detection via a hybrid CNN-Transformer DL framework, specifically working on limitations highlighted in prior research \citep{ref53}. Key contributions include an expanded simulated dataset and an interpretable model architecture. The expanded dataset, with additional simulated PC operational conditions, resulted to be crucial for training models more representative of real-world manufacturing complexities. The increased number of samples, particularly in terms of failure modes, provided richer data to learn subtle fault signatures. This comprehensive dataset enabled a robust evaluation of the procedure using 3 repetitions of 10-fold stratified cross-validation, totaling 30 distinct model training runs. Methodological improvements, notably the strict distinction between training and validation sets before any augmentation or scaling, were critical in mitigating data leakage risks and providing more reliable estimates of the model performance. The use of triplicated 10-fold stratified cross-validation, with different seeds governing data splits and model initializations for each repetition, significantly enhanced the confidence in the reported results and the assessment of model stability.

The TransformerSHM architecture demonstrated exceptionally high classification performance. Current results featuring 99.83\% accuracy, 99.86\% balanced accuracy on a larger, more diverse dataset, do represent a substantial achievement. The Transformer layer, with its attention mechanism, proved to be a key differentiator, effectively capturing inter-sensor dependencies and offering interpretable insights into sensor importance. Sensor importance analysis, derived from the Transformer's attention weights, offered actionable insights for OSP in future PC designs and SHM implementations. Identifying sensors characterized by consistently high-attention and low-variability (like, e.g. sensors 16, 7 and 9) suggested that these locations provide critical diagnostic information in the simulated environment. This knowledge could enable more efficient physical sensor layouts, potentially reducing the number of sensors and network complexity, while maintaining or even improving detection capabilities by focusing on the most informative locations. Experimental validation of these OSP findings represents an important next step.

Despite these significant advancements, certain limitations have to be highlighted:
\begin{itemize}
    \item \textbf{Focus on Specific Failures:} the current research targeted loose screws and cracks only, while real PCs face wider failure arrays. Future work could expand detection capabilities.
    \item \textbf{Simulation-Based Data:} even if expanded, FRF data come from FE simulations under controlled conditions, while real-world data validation is crucial for practical applicability. Domain adaptation may be indeed necessary.
    \item \textbf{Data Augmentation Effects:} Augmentation, while beneficial for dataset expansion and model generalization as evidenced by the learning curves showing no overfitting across all the models, might inadvertently propagate or amplify biases from the original simulated dataset. The physical realism of all augmented samples cannot be definitively guaranteed.
    \item \textbf{Generalizability to Different PC Designs:} The FE model was specifically set for one PC type. Applicability of optimal sensor locations for different PC designs needs investigation.
\end{itemize}

Promising future research avenues include:
\begin{itemize}
   \item \textbf{Hyperparameter and Architecture Optimization:} Implement systematic hyperparameter tuning (e.g., using Bayesian optimization or genetic algorithms) to automatically find the most effective model configuration and architecture, ensuring optimal performance and robustness for this SHM task.
   \item \textbf{Advanced Model Architectures:} explore alternatives like Graph Neural Networks for sensor network topology, or more sophisticated attention mechanisms.
   \item \textbf{Multi-Modal Sensor Fusion:} integrate data from other modalities like touchdown accelerometers, temperature sensors, electrical data, for a comprehensive PC health view. Effective sensor fusion is a key.
    \item \textbf{Real-Time Implementation and Deployment:} develop real-time strategies for manufacturing, considering edge device computation and factory system integration.
    \item \textbf{Enhanced Explainability:} going beyond sensor importance, visualize informative frequency components or temporal patterns for deeper failure mechanism insights.
    \item \textbf{Advanced Data Generation:} exploring techniques like generative adversarial networks for generating synthetic FRF data, with a strong emphasis on ensuring physical realism, to further enhance dataset diversity and robustness.
    \item \textbf{Precise Fault Localization:} investigating the potential of attention weights or similar interpretability techniques for more precise localization of the failure event within the PC.
\end{itemize}

\section{Conclusion}
\label{sec:conclusion}
This research has been aimed at advancing PC failure detection, by leveraging a hybrid DL framework combining CNNs and Transformers with an attention mechanism. Key contributions include the utilization of an expanded and diversified simulated dataset, a rigorous training and evaluation protocol to prevent data leakage, and a novel CNN-Transformer model that achieves outstanding classification accuracy (99.83\% overall and 99.86\% balanced) and provides interpretable sensor importance. Results demonstrated the outstanding performance and robustness of the TransformerSHM model on the simulated data, achieving excellent metrics across all the considered failure classes, including near-perfect recall for critical crack detection (99.7\%) and high precision across all the categories. The attention weight analysis consistently identified crucial sensor contributions, offering a data-driven basis for OSP in future PC SHM designs. This has the potential to lead to more cost-effective and targeted monitoring systems without compromising detection capabilities.

This research has substantiated the significant potential of advanced DL techniques to enhance proactive maintenance strategies in semiconductor manufacturing. The ability to achieve earlier and highly accurate detection of incipient failures in critical components like PCs can substantially improve operational reliability, reduce unscheduled downtime, and ultimately increase manufacturing yield. These findings are pivotal for the development of intelligent, automated systems aimed at elevating the quality and efficiency of semiconductor production. Future work will prioritize experimental validation of these findings on physical systems, and explore pathways for real-world deployment.

\section*{Acknowledgment} 
\label{sec:acknowledgment}
Funded by the European Union (Grant Agreement No. 101072491). Views and opinions expressed are however those of the author(s) only and do not necessarily reflect those of the European Union or the European Research Executive Agency. Neither the European Union nor the granting authority can be held responsible for them.


\begin{thebibliography}{10}

\bibitem{ref1}
T.~H. {Van Nguyen} and C.-F. Chien.
\newblock {Semiconductor probe card proactive maintenance using graph
  self-supervised learning and an empirical study}.
\newblock {\em Computers \& Industrial Engineering}, 203:110955, 2025.
\newblock \doi{10.1016/j.cie.2025.110955}.

\bibitem{ref2}
C.F. Chien and H.J. Wu.
\newblock {Integrated circuit probe card troubleshooting based on rough set
  theory for advanced quality control and an empirical study}.
\newblock {\em Journal of Intelligent Manufacturing}, 35:275--287, 2024.
\newblock \doi{10.1007/S10845-022-02042-8}.

\bibitem{ref3}
W.~Fu, C.F. Chien, and L.~Tang.
\newblock {Bayesian network for integrated circuit testing probe card fault
  diagnosis and troubleshooting to empower Industry 3.5 smart production and an
  empirical study}.
\newblock {\em Journal of Intelligent Manufacturing}, 33(3):785--798, 2022.
\newblock \doi{10.1007/S10845-020-01680-0}.

\bibitem{ref4}
B.~Shin.
\newblock {\em {High-Speed Probe Card Analysis Using Real-time Machine Vision
  and Image Restoration Technique}}.
\newblock Master's Thesis, University of Waterloo, Waterloo, Ontario, Canada,
  2013.

\bibitem{ref5}
C.~F. Chien, Y.~S. Lin, and S.~K. Lin.
\newblock {Deep reinforcement learning for selecting demand forecast models to
  empower Industry 3.5 and an empirical study for a semiconductor component
  distributor}.
\newblock {\em International Journal of Production Research}, 58(9):2784--2804,
  2020.
\newblock \doi{10.1080/00207543.2020.1733125}.

\bibitem{ref6}
J.~M. Nawaz, M.~Z. Arshad, and S.~J. Hong.
\newblock {Fault Diagnosis in Semiconductor Etch Equipment Using Bayesian
  Networks}.
\newblock {\em Journal of Semiconductor Technology and Science}, 14(2):252--261,
  2014.
\newblock \url{https://www.dbpia.co.kr/journal/articleDetail?nodeId=NODE02405138}.

\bibitem{ref7}
B.~Li, T.~Han, and F.~Kang.
\newblock {Fault diagnosis expert system of semiconductor manufacturing
  equipment using a Bayesian network}.
\newblock {\em International Journal of Computer Integrated Manufacturing},
  26(12):1161--1171, 2013.
\newblock \doi{10.1080/0951192X.2013.812803}.

\bibitem{ref8}
H.~Rostami, J.~Blue, and C.~Yugma.
\newblock {Automatic equipment fault fingerprint extraction for the fault
  diagnostic on the batch process data}.
\newblock {\em Applied Soft Computing}, 68:972--989, 2018.
\newblock \doi{10.1016/j.asoc.2017.10.029}.

\bibitem{ref9}
Y.~M. Bae, S.~H. Lee, K.~S. Lee, J.~H. Lim, K.~B. Kim, and B.~H.
  Lee.
\newblock {Detecting abnormal behavior of automatic test equipment using
  autoencoder with event log data}.
\newblock {\em Computers \& Industrial Engineering}, 183:109547, 2023.
\newblock \doi{10.1016/j.cie.2023.109547}.

\bibitem{ref10}
Y.~T. Jou, H.~M. Wee, H.~C. Chen, Y.~H. Hsieh, and L.~Wang.
\newblock {A neural network forecasting model for consumable parts in
  semiconductor manufacturing}.
\newblock {\em Journal of Manufacturing Technology Management}, 20(3):404--412,
  2009.
\newblock \doi{10.1108/17410380910936828}.

\bibitem{ref11}
K.~C.~C. Cheng, C.-J. Su, T.-C. Chang, C.-J. Du, S.~C.
  Hsieh, and Y.-F. Chen.
\newblock {Machine Learning-Based Detection Method for Wafer Test Induced
  Defects}.
\newblock {\em IEEE Transactions on Semiconductor Manufacturing}, 34(2):161--167,
  2021.
\newblock \doi{10.1109/TSM.2021.3065405}.

\bibitem{ref12}
M.~S.~A. Sulaiman, M.~A. Yunus, A.~R. Bahari, and M.~N. Abdul~Rani.
\newblock {Identification of damage based on frequency response function (FRF)
  data}.
\newblock In {\em MATEC Web of Conferences}, volume~90, page 01025, 2016.
\newblock \doi{10.1051/matecconf/20179001025}.

\bibitem{ref13}
R.~P.~B. Kocharla, M.~Kolli, and M.~Cheepu.
\newblock {Real-Time Detection of Faults in Rotating Blades Using Frequency
  Response Function Analysis}.
\newblock {\em Applied Mechanics}, 4(1):356--370, 2023.
\newblock \doi{10.3390/appmech4010020}.

\bibitem{ref14}
J.~E.~T. Penny, M.~I. Friswell, and S.~D. Garvey.
\newblock {Automatic choice of measurement locations for dynamic testing}.
\newblock {\em AIAA journal}, 32(2):407--414, 1994.
\newblock \doi{10.2514/3.11998}.

\bibitem{ref15}
Y.~K. Zhou and Z.~D.
  Duan.
\newblock {Optimum Sensor Placement for Structural Damage Detection}.
\newblock {\em Journal of Engineering Mechanics}, 126(11):1173--1179, 2000.
\newblock \doi{10.1061/(ASCE)0733-9399(2000)126:11(1173)}.

\bibitem{ref16}
K.~Worden and A.~P. Burrows.
\newblock {Optimal sensor placement for fault detection}.
\newblock {\em Engineering Structures}, 23(8):885--901, 2001.
\newblock \doi{10.1016/S0141-0296(00)00118-8}.

\bibitem{ref17}
G.~Capellari, E.~Chatzi, and S.~Mariani.
\newblock {Cost-Benefit Optimization of Structural Health Monitoring Sensor
  Networks}.
\newblock {\em Sensors}, 18(7):2174, 2018.
\newblock \doi{10.3390/S18072174}.

\bibitem{ref18}
G.~Capellari, E.~Chatzi, and S.~Mariani.
\newblock {Structural Health Monitoring Sensor Network Optimization through
  Bayesian Experimental Design}.
\newblock {\em ASCE-ASME Journal of Risk and Uncertainty in Engineering
  Systems, Part A: Civil Engineering}, 4(2):04018016, 2018.
\newblock \doi{10.1061/AJRUA6.0000966}.

\bibitem{ref19}
C.~Malings and M.~Pozzi.
\newblock {Value-of-information in spatio-temporal systems: Sensor placement
  and scheduling}.
\newblock {\em Reliability Engineering \& System Safety}, 172:45--57, 2018.
\newblock \doi{10.1016/j.ress.2017.11.019}.

\bibitem{ref20}
A.~Kamariotis, E.~Chatzi, and D.~Straub.
\newblock {Value of information from vibration-based structural health
  monitoring extracted via Bayesian model updating}.
\newblock {\em Mechanical Systems and Signal Processing}, 166:108465, 2022.
\newblock \doi{10.1016/j.ymssp.2021.108465}.

\bibitem{ref21}
M.~Huang, J.~Li, and H.~Zhu.
\newblock {Optimal sensor layout for bridge health monitoring based on
  dual-structure coding genetic algorithm}.
\newblock In {\em 2009 International Conference on Computational Intelligence
  and Software Engineering (CiSE)}, pages 1--4, 2009.
\newblock \doi{10.1109/CISE.2009.5366481}.

\bibitem{ref22}
T.~H.
  Yi, H.~N. Li, and M.~Gu.
\newblock {Optimal sensor placement for structural health monitoring based on
  multiple optimization strategies}.
\newblock {\em Structural Design of Tall and Special Buildings}, 20(7):881--900,
  2011.
\newblock \doi{10.1002/tal.712}.

\bibitem{ref23}
T.~H. Yi and H.~N. Li.
\newblock {Methodology Developments in Sensor Placement for Health Monitoring
  of Civil Infrastructures}.
\newblock {\em International Journal of Distributed Sensor Networks}, 2012:612726,
  2012.
\newblock \doi{10.1155/2012/612726}.

\bibitem{ref24}
T.~G. Carne and C.~R. Dohrmann.
\newblock {A modal test design strategy for model correlation}.
\newblock Technical Report, Sandia National Labs., Albuquerque, NM, 1994.
\newblock \doi{10.2172/10114604}.

\bibitem{ref25}
J.~W.
  Kim, M.~Torzoni, A.~Corigliano, and S.~Mariani.
\newblock {Attention Mechanism-Driven Sensor Placement Strategy for Structural
  Health Monitoring}.
\newblock {\em Engineering Proceedings}, 27(1):43, 2022.
\newblock \doi{10.3390/ECSA-9-13354}.

\bibitem{ref26}
{ANSYS, Inc.}
\newblock {\em {ANSYS Mechanical, Release R2 2024}}.
\newblock Pittsburgh, PA, 2024.

\bibitem{ref29}
Y.~J.
  Cha, R.~Ali, J.~Lewis, and O.~Büyüköztürk.
\newblock {Deep learning-based structural health monitoring}.
\newblock {\em Automation in Construction}, 161:105328, 2024.
\newblock \doi{10.1016/j.autcon.2024.105328}.

\bibitem{ref27}
J.~Jia and Y.~Li.
\newblock {Deep Learning for Structural Health Monitoring: Data, Algorithms,
  Applications, Challenges, and Trends}.
\newblock {\em Sensors}, 23(21):8824, 2023.
\newblock \doi{10.3390/S23218824}.

\bibitem{ref28}
Y.~Zhang, Q.~Zhou, K.~Zhou, and J.~Tang.
\newblock {Damage Detection of a Pressure Vessel with Smart Sensing and Deep
  Learning}.
\newblock {\em IFAC-PapersOnLine}, 56(3):379--384, 2023.
\newblock \doi{10.1016/j.ifacol.2023.12.053}.

\bibitem{ref30}
C.~Li, J.~Chen, C.~Yang, J.~Yang, Z.~Liu, and P.~Davari.
\newblock {Convolutional Neural Network-Based Transformer Fault Diagnosis
  Using Vibration Signals}.
\newblock {\em Sensors}, 23(10):4781, 2023.
\newblock \doi{10.3390/S23104781}.

\bibitem{ref31}
R.~Zhao, R.~Yan, J.~Wang, and K.~Mao.
\newblock {Learning to Monitor Machine Health with Convolutional Bi-Directional
  LSTM Networks}.
\newblock {\em Sensors}, 17(2):273, 2017.
\newblock \doi{10.3390/S17020273}.

\bibitem{ref32}
J.~Deng, S.~Zhang, and J.~Ma.
\newblock {Self-Attention-Based Deep Convolution LSTM Framework for
  Sensor-Based Badminton Activity Recognition}.
\newblock {\em Sensors}, 23(20):8373, 2023.
\newblock \doi{10.3390/S23208373}.

\bibitem{ref33}
B.~R. Chang, H.~F. Tsai, and Y.~R. Wu.
\newblock {Detection and Prediction of Probe Mark Damage in Wafer Testing}.
\newblock {\em Electronics}, 13(20):4075, 2024.
\newblock \doi{10.3390/electronics13204075}.

\bibitem{ref34}
G.~Toh and J.~Park.
\newblock {Review of Vibration-Based Structural Health Monitoring Using Deep
  Learning}.
\newblock {\em Applied Sciences}, 10(5):1680, 2020.
\newblock \doi{10.3390/app10051680}.

\bibitem{ref35}
S.~Hassani and U.~Dackermann.
\newblock {A Systematic Review of Optimization Algorithms for Structural
  Health Monitoring and Optimal Sensor Placement}.
\newblock {\em Sensors}, 23(6):3293, 2023.
\newblock \doi{10.3390/s23063293}.

\bibitem{ref36}
P.~Konar and P.~Chattopadhyay.
\newblock {Bearing fault detection of induction motor using wavelet and
  Support Vector Machines (SVMs)}.
\newblock {\em Applied Soft Computing}, 11(6):4203--4211, 2011.
\newblock \doi{10.1016/j.asoc.2011.03.014}.

\bibitem{ref37}
A.~Soualhi, K.~Medjaher, and N.~Zerhouni.
\newblock {Bearing health monitoring based on hilbert-huang transform, support
  vector machine, and regression}.
\newblock {\em IEEE Transactions on Instrumentation and Measurement},
  64(1):52--62, 2015.
\newblock \doi{10.1109/TIM.2014.2330494}.

\bibitem{ref38}
A.~C. Neves, L.~Gonz\'{a}lez, J.~Leander, and R.~Karoumi.
\newblock {Structural health monitoring of bridges: a model-free ANN-based
  approach to damage detection}.
\newblock {\em Journal of Civil Structural Health Monitoring}, 7(5):689--702,
  2017.
\newblock \doi{10.1007/s13349-017-0252-5}.

\bibitem{ref39}
R.~A. Osornio-Rios, J.~P. Amezquita-Sanchez, R.~J.
  Romero-Troncoso, and
  A.~Garcia-Perez.
\newblock {MUSIC-ANN Analysis for Locating Structural Damages in a Truss-Type
  Structure by Means of Vibrations}.
\newblock {\em Computer-Aided Civil and Infrastructure Engineering},
  27(9):687--698, 2012.
\newblock \doi{10.1111/j.1467-8667.2012.00777.x}.

\bibitem{ref40}
M.~M. Alamdari, T.~Rakotoarivelo, and N.~L.~D. Khoa.
\newblock {A spectral-based clustering for structural health monitoring of the
  Sydney Harbour Bridge}.
\newblock {\em Mechanical Systems and Signal Processing}, 87:384--400, 2017.
\newblock \doi{10.1016/j.ymssp.2016.10.033}.

\bibitem{ref41}
J.~Tian, M.~H. Azarian, and M.~Pecht.
\newblock {Anomaly Detection Using Self-Organizing Maps-Based K-Nearest
  Neighbor Algorithm}.
\newblock In {\em PHM Society European Conference}, volume~2, 2014.
\newblock \doi{10.36001/phme.2014.v2i1.1554}.

\bibitem{ref42}
O.~Abdeljaber, O.~Avci, S.~Kiranyaz, M.~Gabbouj, and D.~J. Inman.
\newblock {Real-time vibration-based structural damage detection using
  one-dimensional convolutional neural networks}.
\newblock {\em Journal of Sound and Vibration}, 388:154--170, 2017.
\newblock \doi{10.1016/j.jsv.2016.10.043}.

\bibitem{ref43}
M.~A. de~Oliveira, A.~V. Monteiro, and J.~V. Filho.
\newblock {A New Structural Health Monitoring Strategy Based on PZT Sensors
  and Convolutional Neural Network}.
\newblock {\em Sensors}, 18(9):2955, 2018.
\newblock \doi{10.3390/S18092955}.

\bibitem{ref44}
Y.~Z. Lin, Z.~H. Nie, and H.~W. Ma.
\newblock {Structural Damage Detection with Automatic Feature-Extraction
  through Deep Learning}.
\newblock {\em Computer-Aided Civil and Infrastructure Engineering},
  32(12):1025--1046, 2017.
\newblock \doi{10.1111/mice.12313}.

\bibitem{ref45}
H.~D.~M. Onchis.
\newblock {A deep learning approach to condition monitoring of cantilever
  beams via time-frequency extended signatures}.
\newblock {\em Computers in Industry}, 105:177--181, 2019.
\newblock \doi{10.1016/j.compind.2018.12.005}.

\bibitem{ref46}
Y.~Zhao, M.~Noori, W.~A. Altabey, R.~Ghiasi, and Z.~Wu.
\newblock {Deep Learning-Based Damage, Load and Support Identification for a
  Composite Pipeline by Extracting Modal Macro Strains from Dynamic
  Excitations}.
\newblock {\em Applied Sciences}, 8(12):2564, 2018.
\newblock \doi{10.3390/app8122564}.

\bibitem{ref47}
C.~Li, R.~V. Sanchez, G.~Zurita, M.~Cerrada, D.~Cabrera, and R.~E. V\'{a}squez.
\newblock {Gearbox fault diagnosis based on deep random forest fusion of
  acoustic and vibratory signals}.
\newblock {\em Mechanical Systems and Signal Processing}, 76-77:283--293, 2016.
\newblock \doi{10.1016/j.ymssp.2016.02.007}.

\bibitem{ref48}
F.~Yvon.
\newblock {Transformers in Natural Language Processing}.
\newblock In {\em Lecture Notes in Computer Science}, volume 13500 of {\em
  Lecture Notes in Artificial Intelligence}, pages 81--105.
  Springer
  International Publishing, 2023.
\newblock \doi{10.1007/978-3-031-24349-3_6}.

\bibitem{ref49}
A.~Gillioz, J.~Casas, E.~Mugellini, and O.~A. Khaled.
\newblock {Overview of the Transformer-based Models for NLP Tasks}.
\newblock In {\em 2020 15th Conference on Computer Science and Information
  Systems (FedCSIS)}, pages 179--183, 2020.
\newblock \doi{10.15439/2020F20}.

\bibitem{ref50}
J.~Li, W.~Chen, G.~Fan, J.~Li, W.~Chen, and G.~Fan.
\newblock {Structural health monitoring data anomaly detection by transformer
  enhanced densely connected neural networks}.
\newblock {\em Smart Structures and Systems}, 30(6):613--625, 2022.
\newblock \doi{10.12989/sss.2022.30.6.613}.

\bibitem{ref51}
C.~Wang et~al.
\newblock {Learning to Optimise Climate Sensor Placement using a Transformer}.
\newblock {\em arXiv preprint arXiv:2310.12387}, 2023.
\newblock \url{https://arxiv.org/abs/2310.12387}.

\bibitem{ref52}
N.~Fatima, S.~Riaz, S.~Ali, R.~Khan, M.~Ullah, and D.~Kwak.
\newblock {Sensors Faults Classification and Faulty Signals Reconstruction
  Using Deep Learning}.
\newblock {\em IEEE Access}, 12:100544--100558, 2024.
\newblock \doi{10.1109/ACCESS.2024.3425408}.

\bibitem{ref53}
M.~Bejani, D.~Appello, M.~Mauri, E.~Missaglia, and S.~Mariani.
\newblock {Digital Twin-Assisted Optimal Sensor Placement for Real-Time
  Monitoring of Probe Cards in EWS Applications}.
\newblock In {\em 2025 26th International Conference on Thermal, Mechanical
  and Multi-Physics Simulation and Experiments in Microelectronics and
  Microsystems (EuroSimE)}, pages 1--6, 2025.
\newblock \doi{10.1109/EUROSIME65125.2025.11006625}.

\bibitem{ref54}
S.~Karmakov and M.~H. Ferri~Aliabadi.
\newblock {Deep Learning Approach to Impact Classification in Sensorized
  Panels Using Self-Attention}.
\newblock {\em Sensors}, 22(12):4370, 2022.
\newblock \doi{10.3390/S22124370}.

\bibitem{ref55}
X.~Zhang, P.~Li, X.~Han, Y.~Yang, and Y.~Cui.
\newblock {Enhancing Time Series Product Demand Forecasting With Hybrid
  Attention-Based Deep Learning Models}.
\newblock {\em IEEE Access}, 12:190079--190091, 2024.
\newblock \doi{10.1109/ACCESS.2024.3516697}.

\bibitem{Rosafalco2020}
L.~Rosafalco, A.~Manzoni, S.~Mariani, and A.~Corigliano.
\newblock {Fully convolutional networks for structural health monitoring
  through multivariate time series classification}.
\newblock {\em Advanced Modeling and Simulation in Engineering Sciences},
  7(1):38, 2020.
\newblock \doi{10.1186/s40323-020-00174-1}.

\end{thebibliography}
\end{document}